\pdfoutput=1

\documentclass[11pt]{article}

\usepackage[]{latex/acl}

\usepackage{times}
\usepackage{latexsym}

\usepackage[T1]{fontenc}

\usepackage[utf8]{inputenc}

\usepackage{microtype}

\usepackage{inconsolata}
\usepackage{multirow}
\usepackage{subcaption}
\usepackage{graphicx}
\usepackage{amssymb}
\usepackage{amsthm}
\usepackage{amsmath,amssymb,amsfonts}
\usepackage{booktabs}
\usepackage{pifont}
\usepackage[ruled, linesnumbered, vlined]{algorithm2e}
\usepackage{verbatim}
\SetKwInput{KwInput}{Input}                
\SetKwInput{KwOutput}{Output}
\SetKwProg{Fn}{Procedure}{}{end}
\colorlet{Red}{gray!40!red}
\colorlet{Green}{gray!40!green}
\newtheorem{definition}{Definition}
\title{Temporal Fact Reasoning over Hyper-Relational Knowledge Graphs}

\author{Zifeng Ding\thanks{Equal contribution. Work done at LMU Munich.}$^{1,2,3}$, Jingcheng Wu\footnotemark[1]$^{4}$, Jingpei Wu$^{1}$, Yan Xia$^{3,5}$, \\ \textbf{Bo Xiong\thanks{Corresponding author.}$^{4}$, Volker Tresp\footnotemark[2]$^{1}$} \\
$^{1}$LMU Munich
$^{2}$University of Cambridge\\ $^{3}$Munich Center for Machine Learning (MCML) $^{4}$University of Stuttgart\\
$^{5}$Technical University of Munich\\
\texttt{zd320@cam.ac.uk,}\\
\texttt{\{jingpei.wu,tresp\}@dbs.ifi.lmu.de,}\\
\texttt{\{bo.xiong, jingcheng.wu\}@ki.uni-stuttgart.de,}\\
\texttt{yan.xia@tum.de}\\}

\begin{document}
\maketitle
\begin{abstract}
Stemming from traditional knowledge graphs (KGs), hyper-relational KGs (HKGs) provide additional key-value pairs (i.e., qualifiers) for each KG fact that help to better restrict the fact validity. In recent years, there has been an increasing interest in studying graph reasoning over HKGs. 
Meanwhile, as discussed in recent works that focus on temporal KGs (TKGs), world knowledge is ever-evolving, making it important to reason over temporal facts in KGs.
Previous mainstream benchmark HKGs do not explicitly specify temporal information for each HKG fact. Therefore, almost all existing HKG reasoning approaches do not devise any module specifically for temporal reasoning. To better study temporal fact reasoning over HKGs, we propose a new type of data structure named hyper-relational TKG (HTKG). Every fact in an HTKG is coupled with a timestamp explicitly indicating its time validity. 
We develop two new benchmark HTKG datasets, i.e., Wiki-hy and YAGO-hy, and propose an HTKG reasoning model that efficiently models hyper-relational temporal facts. To support future research on this topic, we open-source our datasets and model\footnote{https://github.com/0sidewalkenforcer0/HypeTKG}.
\end{abstract}

\section{Introduction}
\begin{figure}[t!]
    \centering
    \includegraphics[width=0.9\columnwidth]{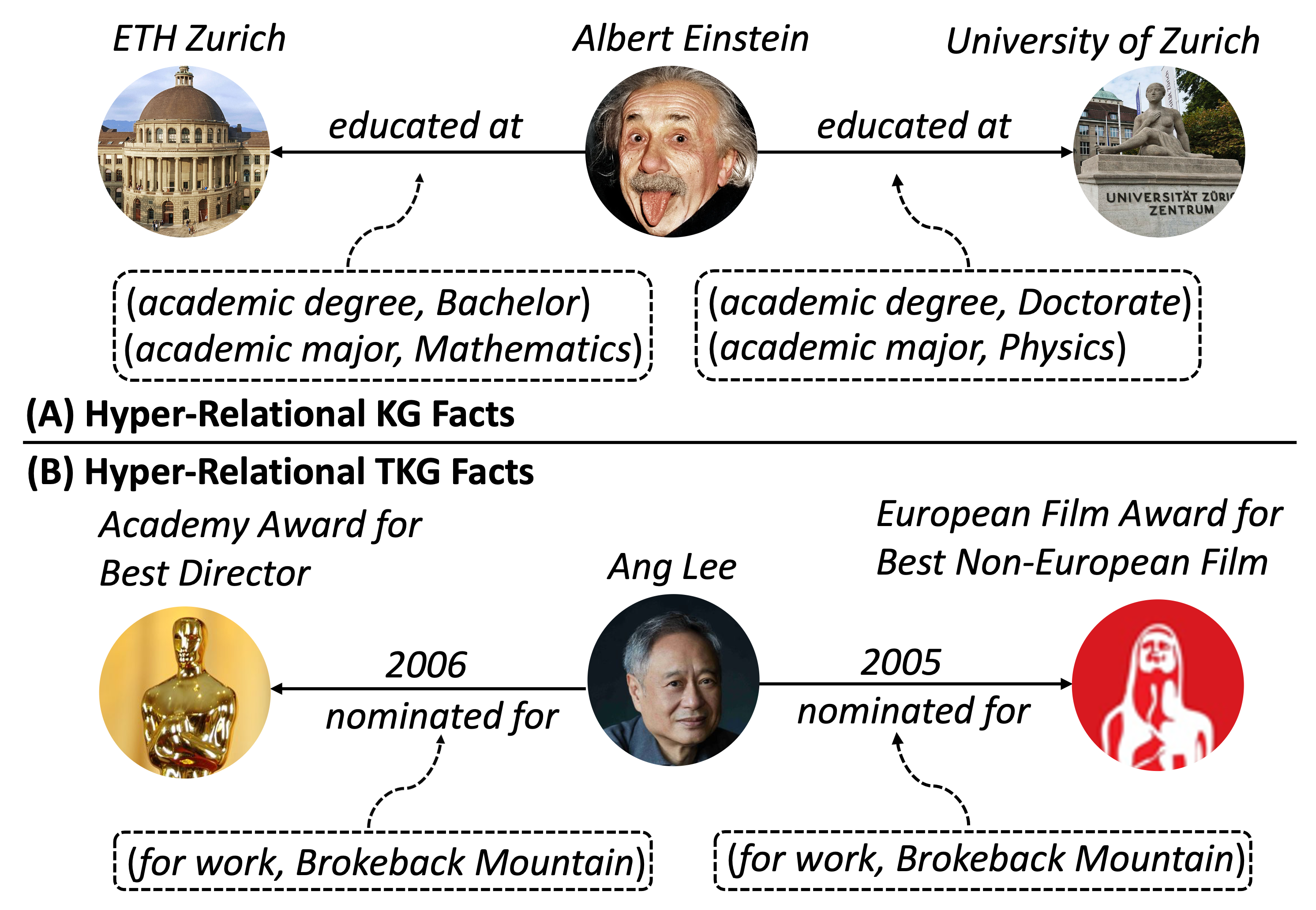}
    \caption{Examples of HKG (A) and HTKG (B) facts. Contents inside dashed line squares denote qualifiers. We also provide another example of HTKG fact showcasing diverse sets of qualifiers in App. \ref{app: more htkg example}.}
    \label{fig: hyper example}
\end{figure}

Traditional knowledge graphs (KGs) represent world knowledge by storing a collection of facts in the form of triples. Each KG fact can be described as $(s, r, o)$, where $s$, $o$ are the subject and object entities of the fact and $r$ denotes the relation between them. 
On top of traditional triple-based KGs, hyper-relational KGs (HKGs) are designed to introduce additional information into each triple-based fact (also known as primary triple in HKGs) by incorporating a number of key-value restrictions named as qualifiers \cite{DBLP:conf/www/ZhangLMM18,DBLP:conf/www/GuanJWC19,DBLP:conf/emnlp/GalkinTMUL20}. Compared with triple-based KGs, HKGs provide more complicated semantics. For example, in Fig. \ref{fig: hyper example} (A), the degree and major information of \textit{Albert Einstein} is provided, which helps to differentiate between the facts regarding two universities attended by him.


Many reasoning approaches have been proposed for HKGs, e.g., \cite{DBLP:conf/acl/WangWLZ21,DBLP:conf/acl/XiongNPS23}, but unfortunately, they all assume that the hyper-relational facts are static. 
As discussed in recent works \cite{DBLP:conf/emnlp/DasguptaRT18,ding2022a}, world knowledge is ever-evolving. In temporal KGs, each fact is represented by a quadruple $(s,r,o,t)$ with an additional timestamp specifying the time validity. Previous mainstream HKG benchmarks do not explicitly specify time validity for each HKG fact. This hinders the development of the reasoning systems that can effectively handle temporal dynamics within hyper-relational facts, and as a result, almost all existing HKG reasoning methods lack a dedicated module for temporal reasoning.
Modeling temporal knowledge in HKGs is important as the temporal validity of a fact improves knowledge expressiveness and might be correlative to its qualifiers. A model should be expressive enough to model such correlation.



To better study temporal fact reasoning over HKGs, we propose a new type of data structure named hyper-relational TKG (HTKG, see formal definition in Sec. \ref{sec:definition}). Every fact in an HTKG is defined in the form of $((s,r,o,t), \{(r_{q_i}, e_{q_i})\}_{i=1}^n)$. $(s,r,o,t)$ is its primary quadruple (i.e., primary fact, $t$ is a timestamp denoting the valid time)
and $\{(r_{q_i}, e_{q_i})\}_{i=1}^n$ are a number of $n$ augmented qualifiers.
We illustrate an HTKG fact example in Fig. \ref{fig: hyper example} (B).
The two awards \textit{Ang Lee} was nominated for because of \textit{Brokeback Mountain} can be differentiated considering the specified timestamps.
An HTKG is composed solely of a collection of hyper-relational temporal facts so we use HTKGs to study temporal fact reasoning over HKGs.
We construct two benchmark HTKGs Wiki-hy and YAGO-hy based on two traditional TKG benchmarks Wikidata11k \cite{DBLP:conf/kdd/JungJK21} and YAGO1830 \cite{DBLP:conf/iclr/HanCMT21}. 

Since previous HKG reasoning approaches pay little attention to temporal reasoning, they are not fit for modeling HTKGs. To this end, we develop a model to achieve link prediction (LP) over \textbf{hype}r-relational \textbf{TKG}s (HypeTKG) as follows: 
(1) We first devise a qualifier-attentional time-aware graph encoder (QATGE) that
considers both temporal information and qualifiers in the graph aggregation process.
(2) We then design a qualifier matching decoder (QMD). Given any HTKG LP query, QMD not only considers its own qualifiers, 
but also models all the qualifiers appearing in query subject-related facts\footnote{Each query subject-related fact is a fact that takes the query subject as the subject entity of its primary quadruple.}.
The motivation of QMD is that the evidence for LP not only is stored in the query qualifiers but also can be found in other subject-related facts. 
Compared with previous methods, HypeTKG is able to capture the correlation between temporal validity and qualifiers.

Another point worth noting is that some recent works have started to explore whether time-invariant (TI) relational knowledge\footnote{TI knowledge are represented with fact triples $(s,r,o)$ (same as the facts in triple-based KGs) and are valid anytime.} can help to enhance temporal fact reasoning on traditional TKGs \cite{DBLP:conf/sigir/LiJLGGSWC21,DBLP:conf/ijcai/LiS022,DBLP:conf/icde/Liu0X0023}. This arouses our interest in studying whether TI relational facts are beneficial in HTKG reasoning. In our work, we mine the TI relational knowledge from the Wikidata KB. 
We pick out the facts that contain ten frequently mentioned TI relations, e.g., \textit{official language}, and ensure that these facts remain valid within the whole time scopes of HTKGs. We adjust HypeTKG and create a model variant HypeTKG$^\psi$ that dynamically controls the influence of TI information for better reasoning on temporal facts. We also provide a wide range of baselines with TI facts and benchmark their temporal fact LP performance on our proposed HTKGs.

To summarize, our contribution is three-folded: (1) We propose a new data structure HTKG that draws attention to temporal fact reasoning over HKGs and propose two corresponding benchmarks (Sec. \ref{sec:definition} and \ref{sec: benchmark}). (2) We propose HypeTKG, a model specifically designed to reason over HTKGs. Experimental results show that HypeTKG performs well in temporal fact reasoning over HTKGs (Sec. \ref{sec: main results}). (3) We study the influence of TI relational knowledge on HTKG reasoning and adapt HypeTKG to accommodate to TI information. We show that our model can benefit by carefully balancing the information between temporal and TI knowledge (Sec. \ref{sec: static result}).

\section{Preliminaries and Related Work}
\subsection{Definition and Problem Statement}
\label{sec:definition}
\begin{definition}[Hyper-Relational TKG]
Let $\mathcal{E}$, $\mathcal{R}$, $\mathcal{T}$ denote a set of entities, relations and timestamps\footnote{We decompose time periods into a series of timestamps following \cite{DBLP:conf/emnlp/JinQJR20}.}, respectively. An HTKG $\mathcal{G}$ is a set of hyper-relational temporal facts. Each fact is denoted as $((s,r,o,t), \{(r_{q_i}, e_{q_i})\}_{i=1}^n)$, where $(s,r,o,t)$ is its primary quadruple. $e_{q_i} \in \mathcal{E}$ and $r_{q_i} \in \mathcal{R}$ are the entity and relation in its $i^{\text{th}}$ qualifier $q_i$, respectively. $n$ is the number of qualifiers.
\end{definition} 
\begin{definition}[Hyper-Relational TKG LP]
    Let $\mathcal{G}_\text{tr}$ be a ground-truth HTKG. $\mathcal{G}_\text{tr} = \mathcal{G}_\text{obs} \cup \mathcal{G}_\text{un}$ ($\mathcal{G}_\text{obs} \cap \mathcal{G}_\text{un} = \emptyset$), where $\mathcal{G}_\text{obs}$ is a set of observed HTKG facts and $\mathcal{G}_\text{un}$ is a set of unobserved facts. Given $\mathcal{G}_\text{obs}$, HTKG LP aims to predict the missing entity in the LP query $\left(\left(s,r,?,t\right), \{(r_{q_i}, e_{q_i})\}_{i=1}^n\right)$ (or $((?,r,o,t), \{(r_{q_i}, e_{q_i})\}_{i=1}^n)$) derived from each fact in $\mathcal{G}_\text{un}$.
\end{definition}

Following previous works on TKGs, e.g., \cite{DBLP:conf/emnlp/HanDMGT21}, for each fact, we create another fact $((o,r^{-1},s,t), \{(r_{q_i}, e_{q_i})\}_{i=1}^n)$ and add it to the graph, where $r^{-1}$ denotes $r$'s inverse relation. We derive an object entity prediction query from each fact and perform object prediction.
Note that we follow \cite{DBLP:conf/emnlp/GalkinTMUL20} and only predict missing entities in primary facts.
\subsection{Related Work}
Due to page limit, see App. \ref{app: related work} for the detailed discussion of various previous methods.
\paragraph{Temporal Fact Reasoning on Traditional TKGs}
Extensive research has been conducted for TKG reasoning. Although traditional TKG facts have no qualifiers, each of them has a specified time identifier for temporal fact reasoning.
A series of works develops time-aware score functions \cite{DBLP:conf/www/LeblayC18,DBLP:conf/coling/XuNAYL20,DBLP:conf/aaai/GoelKBP20,DBLP:journals/kbs/ShaoZYTCL22,DBLP:conf/aaai/MessnerAC22,DBLP:conf/acl/LiSG23,DBLP:conf/aaai/PanNLS24} that compute plausibility scores of quadruple-based TKG facts based on various types of geometric operations. Some other methods employ neural structures, e.g., LSTM \cite{article} or time-aware graph neural networks, to achieve temporal reasoning \cite{DBLP:conf/emnlp/JinQJR20,DBLP:conf/emnlp/WuCCH20,DBLP:conf/emnlp/HanDMGT21,DBLP:conf/aaai/ZhuCFCZ21,DBLP:conf/sigir/LiJLGGSWC21,DBLP:conf/kdd/JungJK21,ding2022a,DBLP:conf/ijcai/LiS022,DBLP:conf/icde/Liu0X0023,DBLP:conf/naacl/DingCWMLXT24}. There are two settings in TKG LP, i.e., interpolation and extrapolation. In extrapolation, to predict a fact happening at time $t$, models can only observe previous TKG facts before $t$, while such restriction is not imposed in interpolation. 
In our work, we only focus on the interpolated LP on HTKGs and leave extrapolation for future work.
\paragraph{Hyper-Relational KG Reasoning}
Mainstream HKG reasoning methods can be categorized into three types.
The first type of works \cite{DBLP:conf/www/ZhangLMM18,DBLP:conf/www/0016Y020,DBLP:conf/ijcai/FatemiTV020,DBLP:conf/www/DiYC21,DBLP:conf/www/WangWLC023} treats each hyper-relational fact as an $n$-ary fact represented with an $n$-tuple: $r_\text{abs}(e_1,e_2, ..., e_n)$, where $n$ is the non-negative arity of an abstract relation $r_\text{abs}$\footnote{Abstract relation $r_\text{abs}$ is derived from a combination of several KG relations by concatenating the relations in the primary triple and qualifiers \cite{DBLP:conf/emnlp/GalkinTMUL20}.} representing the number of entities involved within $r_\text{abs}$ and $e_1, ..., e_n$ are the entities appearing in this $n$-ary fact. 
Although these methods show strong effectiveness, previous study \cite{DBLP:conf/emnlp/GalkinTMUL20} has shown that the way of treating HKG facts as $n$-ary facts naturally loses the semantics of the original KG relations and would lead to a combinatorial explosion of relation types.
The second type of works \cite{DBLP:conf/www/LiuYL21,DBLP:journals/tkde/GuanJGWC23} transforms each hyper-relational fact into a set of key-value pairs: $\{(r_i:e_i)\}_{i=1}^n$. 
Formulating hyper-relational facts into solely key-value pairs would also cause a problem that the relations from the primary fact triples and qualifiers cannot be fully distinguished \cite{DBLP:conf/emnlp/GalkinTMUL20}. 
To overcome the problems incurred in first two types of methods, recently, some works \cite{DBLP:conf/acl/GuanJGWC20,DBLP:conf/www/RossoYC20,DBLP:conf/emnlp/GalkinTMUL20,DBLP:conf/acl/WangWLZ21,DBLP:conf/acl/XiongNPS23,DBLP:conf/kdd/ChungLW23} formulate each hyper-relational fact into a primary triple with a set of key-value qualifier pairs: $\{((s,r,o),\{(r_{q_i}, e_{q_i})\}_{i=1}^n)\}$. This formulation distinguishes the primary fact triples and qualifiers, and meanwhile preserves the semantics of the original KG relations. While HKG reasoning methods perform well on HKG LP, none of them focuses on temporal reasoning because no temporal identifiers are explicitly specified in HKGs.
To draw attention to temporal fact reasoning over hyper-relational facts,
a recent work \cite{DBLP:conf/emnlp/HouJL0GZGC23} proposes n-tuple TKG (N-TKG), where each hyper-relational fact is represented with an n-tuple: $(r, \{\rho_i: e_i\}_{i=1}^n, t)$. $n$ and $t$ are the arity and the timestamp of the fact, respectively. $\rho_i$ is the labeled role of the entity $e_i$. $r$ denotes fact type. Compared with HTKG, N-TKG has limitation: HTKGs explicitly separate primary facts with additional qualifiers, while N-TKGs mix all the entities from the primary facts and qualifiers and are unable to fully emphasize the importance of primary facts. Hou et al. also propose a model NE-Net for extrapolated LP on N-TKGs. It is not optimal for interpolation because it can only model the graph information before the prediction timestamp. See App. \ref{app: related work} for more discussion.
\begin{table*}[t]
    \centering
    \resizebox{\textwidth}{!}{
\begin{tabular}{c c c c c c c c c c c c c c}
\toprule
Dataset&$N_\textrm{train}$&$N_\textrm{valid}$&$N_\textrm{test}$&$|\mathcal{E}_\text{pri}|$&$|\mathcal{E}_{\text{Qual}}|$&$|\mathcal{R}_{\text{pri}}|$&$|\mathcal{R}_{\text{Qual}}|$&$|\mathcal{T}|$ & $|\exists$ Qual$|$ & avg$(|\text{Qual}|)$ & Qual\% & $|\mathcal{G}_\text{TI}|$&$|\mathcal{E}_\text{TI}|$\\ 
\midrule
Wiki-hy  & $111,252$ & $13,900$ & $13,926$ & $11,140$ & $1,642$ & $92$ & $44$ & $508$ & $26,670$ & $1.59$ & $9.59\%$ & $5,360$& $3,801$\\ 
YAGO-hy  & $51,193$ & $10,973$ & $10,977$ & $10,026$ & $359$ & $10$ & $33$ & $188$ & $10,214$ & $1.10$ & $6.98\%$ & $7,331$& $5,782$\\ 
\bottomrule
\end{tabular}
}
\caption{Dataset statistics. $N_\textrm{train}$/$N_\textrm{valid}$/$N_\textrm{test}$ is the number of facts in the training/validation/test set. $|\mathcal{E}_\text{pri}|/|\mathcal{R}_\text{pri}|/|\mathcal{T}|$ is the number of entities/relations/timestamps in primary quadruples.  $|\mathcal{E}_\text{Qual}|/|\mathcal{R}_\text{Qual}|$ is the number of additional entities/relations only existing in qualifiers. $|\exists$ Qual$|$/Qual\% is the number/the proportion of facts containing at least one qualifier.
Complete sets of entities and relations are $\mathcal{E} = \mathcal{E}_\text{pri} \cup \mathcal{E}_\text{Qual}$ and $\mathcal{R} = \mathcal{R}_\text{pri} \cup \mathcal{R}_\text{Qual}$, respectively. $\mathcal{E}_\text{TI}$ is the number of entities additionally introduced in TI facts $\mathcal{G}_\text{TI}$ and $\mathcal{E}_\text{TI} \cap \mathcal{E} = \emptyset$.}
\label{tab: data}
\end{table*}
\section{Proposing New Benchmarks}
\label{sec: benchmark}
We propose two HTKG benchmark datasets Wiki-hy and YAGO-hy. Wiki-hy contains HTKG facts extracted from Wikidata \cite{DBLP:journals/cacm/VrandecicK14}, where they happen from year $1513$ to $2020$. YAGO-hy is constructed from the facts in YAGO3 \cite{DBLP:conf/cidr/MahdisoltaniBS15} and the time scope is from year $1830$ to $2018$. We use previous traditional TKG benchmarks Wikidata11k \cite{DBLP:conf/kdd/JungJK21} and YAGO1830 \cite{DBLP:conf/iclr/HanCMT21} as bases and search for the qualifiers of their facts in Wikidata. We use the MediaWiki API\footnote{https://www.wikidata.org/w/api.php} to identify the quadruple-based TKG facts in Wikidata and extract all the qualifiers stated under the corresponding Wikidata statements. Since Wikidata11k is originally extracted from Wikidata, we can directly find its relations and entities in this KB. YAGO1830's entities share the same pool as Wikidata but relation types are taken from \textit{schema.org}. We map YAGO1830's relations to Wikidata's relations to enable fact matching (detailed mapping in App. \ref{app: yago mapping}). 
We provide dataset statistics of both datasets in Table \ref{tab: data}. Qualifier searching will include additional entities and relations. We include them in model training and evaluation. We augment quadruple-based TKG facts with their searched qualifiers. The facts without any searched qualifier will remain unchanged.
All the facts in our datasets are based on English.
We discuss why we use Wikidata-based but not other popular ICEWS-based TKGs to construct HTKGs in App. \ref{app: why not icews}.
\begin{figure*} 
    \centering
  \subfloat[Qualifier-attentional time-aware graph encoder (QATGE).\label{fig: encoder}]{%
       \includegraphics[width=0.49\textwidth]{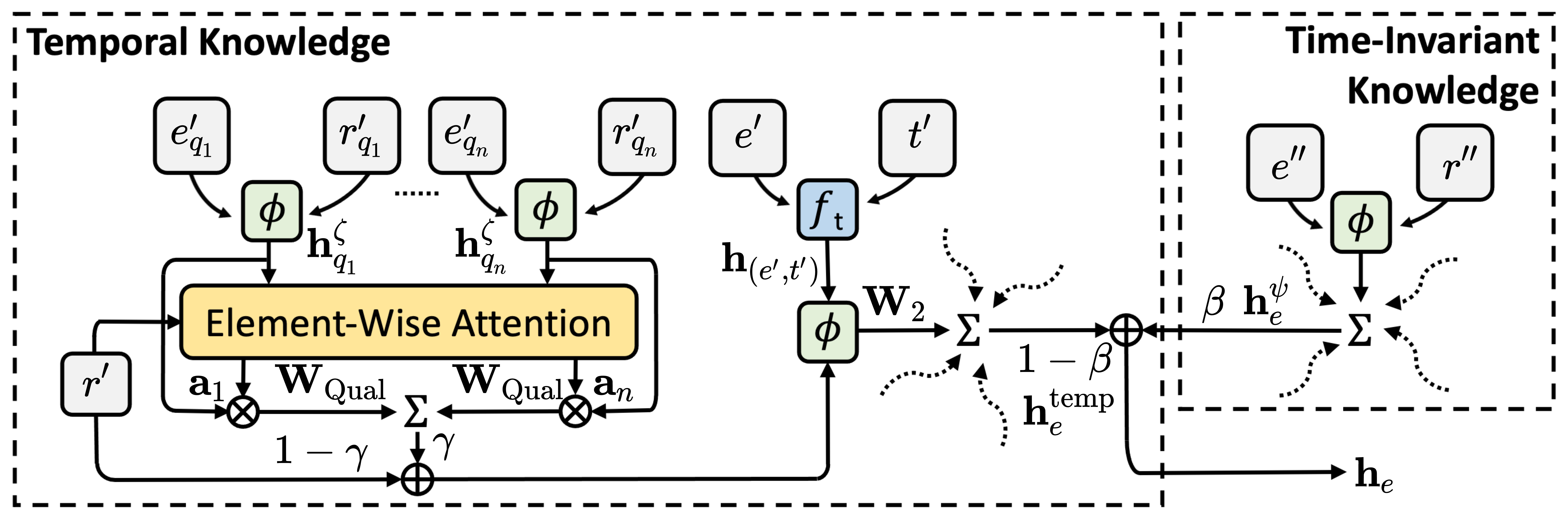}}
    \hfill
  \subfloat[Qualifier matching decoder (QMD).\label{fig: decoder}]{%
        \includegraphics[width=0.49\textwidth]{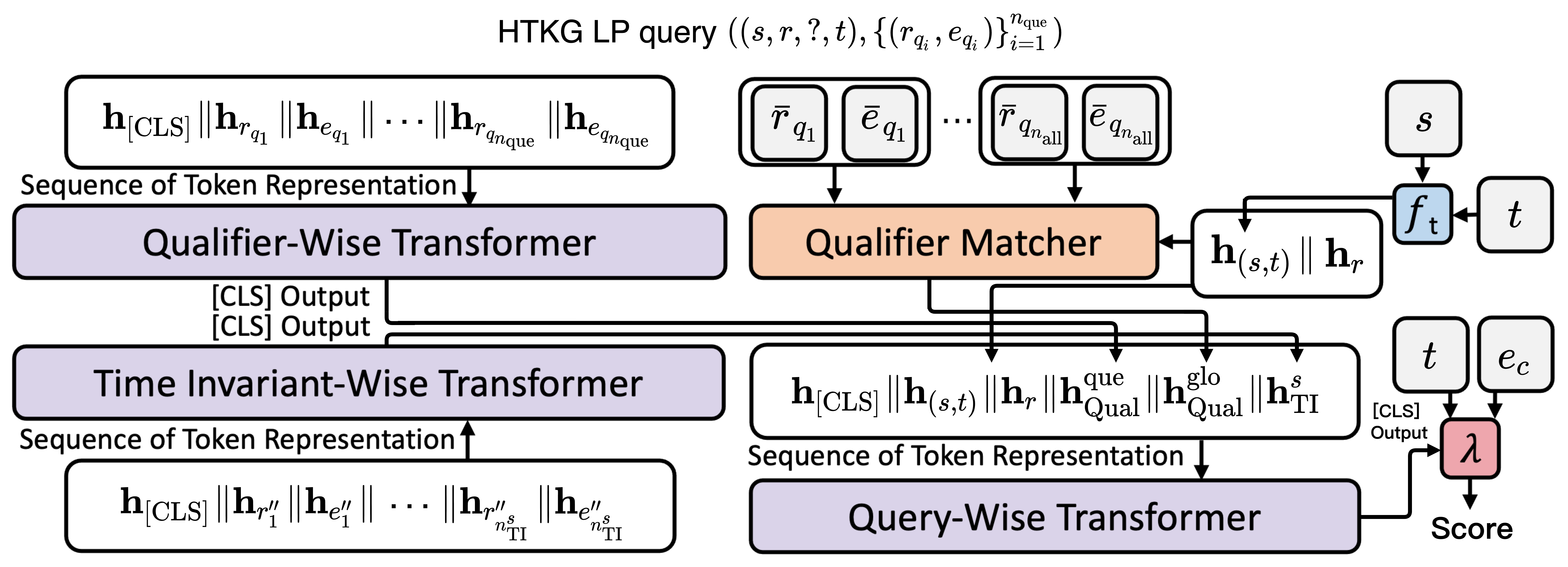}}
    \hfill
  \caption{Model structure of HypeTKG$^\psi$. HypeTKG$^\psi$ first uses QATGE to encode all the entities. It then uses QMD to compute score regarding every candidate entity $e_c \in \mathcal{E}$.
  Temporal information is considered in both QATGE and QMD for temporal reasoning. The structure of HypeTKG can be derived by excluding the components concerning TI facts. View with Sec. \ref{sec: model} for better understanding. $e''_1, ..., e''_{n_\text{TI}^s}$ and $r''_1, ..., r''_{n_\text{TI}^s}$ are the entities and relations from a number of $n_\text{TI}^s$ TI neighbors of query subject $s$, respectively. 
  }
  \label{fig: model structure} 
\end{figure*}

We explore TI knowledge as follows. We first find the top 400 frequent relations in Wikidata KB. Based on them, we then manually check each of them and pick out top 10 frequent relations that describe TI relationships among entities. The selected TI relations are {\textit{family name}, \textit{native language}, \textit{subclass of}, \textit{official language}, \textit{child}, \textit{sibling}, \textit{father}, \textit{mother}, \textit{ethnic group}, \textit{country of origin}}. We ensure that they are disjoint from the existing relations in the original HTKGs. Starting from the entities in our HTKGs, we search for their associated TI facts in Wikidata, where each of them corresponds to a selected TI relation. For example, for the YAGO-hy entity \textit{Emmy Award}, we take the facts such as (\textit{Emmy Award}, \textit{subclass of}, \textit{television award}). 
As a result, we collect a set of facts denoted as $\mathcal{G}_\text{TI}$ ($\mathcal{G}_\text{TI} \cap \mathcal{G}_\text{tr}=\emptyset$) for Wiki-hy and YAGO-hy. We allow models to use all of them for enhancing LP over temporal facts during train/valid/test. See Table \ref{tab: data} for $\mathcal{G}_\text{TI}$ statistics. 
\section{HypeTKG}
\label{sec: model}
HypeTKG consists of two parts, i.e., a qualifier-attentional time-aware graph encoder (QATGE) and a qualifier matching decoder (QMD). To further learn from TI knowledge, we equip HypeTKG with additional modules and develop a model variant HypeTKG$^\psi$ (model structure shown in Fig. \ref{fig: model structure}). 
\subsection{Qualifier-Attentional Time-Aware Graph Encoder}
\label{sec: encoder}
QATGE learns a contextualized representation for every entity. Given an entity $e$, QATGE first finds its temporal neighbors from $\mathcal{G}_\text{obs}$: $\mathcal{N}_{e} = \{\zeta\} = \{((e',r',t'), \{(r'_{q_i}, e'_{q_i})\}_{i=1}^n)\}$, where each temporal neighbor $\zeta$ is derived from a fact $((e',r', e, t'), \{(r'_{q_i}, e'_{q_i})\}_{i=1}^n) \in \mathcal{G}_\text{obs}$ connecting to $e$.
For each $\zeta$, QATGE employs an attention-based module to model its qualifiers. It computes the representation $\mathbf{h}^\zeta_{q_i}$ for the $i^\text{th}$ qualifier $q_i$ of $\zeta$ with a function $\phi(\cdot,\cdot)$.
\begin{equation}
\small
\resizebox{0.91\columnwidth}{!}{%
$
\begin{aligned}
    \mathbf{h}^{\zeta}_{q_i} &= \phi (\mathbf{h}_{e'_{q_i}}, \mathbf{h}_{r'_{q_i}}) \\
    &=  \mathbf{W}_1 (\mathbf{h}_{e'_{q_i}} \| \mathbf{h}_{r'_{q_i}}) * f(\mathbf{h}^\mathbb{C}_{e'_{q_i}} \circ  \mathbf{h}^\mathbb{C}_{r'_{q_i}}) * (\mathbf{h}_{e'_{q_i}} \oplus \mathbf{h}_{r'_{q_i}}).
\end{aligned}
$}
\end{equation}
$\mathbf{h}_{e'_{q_i}} \in \mathbb{R}^d$ and $\mathbf{h}_{r'_{q_i}} \in \mathbb{R}^d$ denote the representations of the entity and relation in $q_i$, respectively. $\|$ means concatenation and $\mathbf{W}_1 \in \mathbb{R}^{d\times2d}$ is a weight matrix. $\mathbf{h}^\mathbb{C}_{e'_{q_i}}  \in \mathbb{C}^{\frac{d}{2}}$ and $\mathbf{h}^\mathbb{C}_{r'_{q_i}} \in \mathbb{C}^{\frac{d}{2}}$ are the complex vectors mapped from $\mathbf{h}_{e'_{q_i}}$ and $\mathbf{h}_{r'_{q_i}}$. The real part of $\mathbf{h}^\mathbb{C}_{e'_{q_i}}$ is the first half of $\mathbf{h}_{e'_{q_i}}$ and the imaginary part is the second half (see mapping explanation and example in App. \ref{app: complex vector}).
$\circ$ is the Hadmard product on the complex space. $f(\cdot): \mathbb{C}^{\frac{d}{2}} \rightarrow \mathbb{R}^{d}$ is a mapping function that maps the complex vectors back to the real vectors. 
$*$ and $\oplus$ are element-wise product and add operations, respectively. After getting $\{\mathbf{h}^\zeta_{q_i}\}$, QATGE integrates the information from all of them by computing an attentional feature $\mathbf{h}^\zeta_\text{Qual}$ related to the primary relation $r'$ of $\zeta$.
\begin{equation}
\small
\resizebox{0.9\columnwidth}{!}{%
$
\begin{aligned}
&\Tilde{\mathbf{h}}^\zeta_{q_i} = ({\mathbf{h}^\zeta_{q_i}}^\top \mathbf{h}_{r'}) * \mathbf{w},\\
&\alpha_i[j] = \frac{\text{exp}(\Tilde{\mathbf{h}}^\zeta_{q_i}[j])}{\sum_{k=1}^n \text{exp}(\Tilde{\mathbf{h}}^\zeta_{q_k}[j])}; \ \mathbf{a}_i = [\alpha_i[1], ..., \alpha_i[d]]^\top,\\
&\mathbf{h}^\zeta_\text{Qual} = \sum_{q_i} \mathbf{W}_\text{Qual} (\mathbf{a}_i * \mathbf{h}^\zeta_{q_i}).
\end{aligned}
$}
\end{equation}
$\mathbf{w} \in \mathbb{R}^d$ is a trainable parameter. 
$\Tilde{\mathbf{h}}^\zeta_{q_i}[j]$ denotes the $j^\text{th}$ element of $\Tilde{\mathbf{h}}^\zeta_{q_i}$.
$\mathbf{a}_i$ is an attention vector, where each of its element $\alpha_i[j]$ denotes the attention score determining how important the $j^\text{th}$ element of the $i^\text{th}$ qualifier $q_i$ is in the $j^\text{th}$ element of $\mathbf{h}^\zeta_\text{Qual}$. The importance increases as the score rises. $\mathbf{W}_\text{Qual} \in \mathbb{R}^{d\times d}$ is a weight matrix. $\mathbf{h}^\zeta_\text{Qual}$ can be viewed as a parameter that adaptively selects the information highly-related to $r'$ from all the qualifiers of $\zeta$. To compute $e$'s representation $\mathbf{h}_e$, we aggregate over all its temporal neighbors in $\mathcal{N}_e$ with a gated structure.
\begin{equation}
\label{eq: QATGE}
\small
\resizebox{\columnwidth}{!}{%
$
    \mathbf{h}_e = \frac{1}{|\mathcal{N}_e|}\sum_{\zeta \in \mathcal{N}_e} \mathbf{W}_2 \phi\left(\mathbf{h}_{(e', t')}, \left(\gamma \mathbf{h}^\zeta_\text{Qual} + (1-\gamma)\mathbf{h}_{r'}\right)\right),
$}
\end{equation}
where $\mathbf{W}_2 \in \mathbb{R}^{d\times d}$ is a weight matrix. $\gamma$ is a trainable gate parameter controlling the amount of information taken from either the primary relation $r'$ or the qualifiers. QATGE incorporates temporal information by learning a time-aware representation for each temporal neighbor's subject entity: $\mathbf{h}_{(e', t')} = f_t(\mathbf{h}_{e'}\|\mathbf{h}_{t'})$. $f_t(\cdot):\mathbb{R}^{2d} \rightarrow \mathbb{R}^d$ is a layer of neural network. $\mathbf{h}_{t'} = \sqrt{1/d}[cos(\omega_1t'+\phi_1), \dots, cos(\omega_{d}t'+ \phi_{d})]$, where $\omega_1 \dots \omega_{d}$ and $\phi_1 \dots \phi_{d}$ are trainable parameters.
\subsection{Qualifier Matching Decoder}
\label{sec: decoder}
QMD leverages the entity and relation representations encoded by QATGE for LP. Assume we want to predict the missing entity of the LP query $((s,r,?,t), \{(r_{q_i}, e_{q_i})\}_{i=1}^{n_\text{que}})$ ($n_\text{que}$ is the number of query qualifiers), 
QMD learns a query feature $\mathbf{h}^\text{que}$. QMD first models query qualifiers $\{(r_{q_i}, e_{q_i})\}_{i=1}^{n_\text{que}}$ with a qualifier-wise Transformer \cite{DBLP:conf/nips/VaswaniSPUJGKP17}. Each query qualifier's entity and relation are treated as two tokens and concatenated as a sub-sequence for this qualifier.
The classification ([CLS]) token is then concatenated with the query qualifier tokens as a sequence and input into the qualifier-wise Transformer, where the sequence length is $2n_\text{que}+1$. We take the output representation of the [CLS] token as the query qualifier feature $\mathbf{h}^\text{que}_\text{Qual} \in \mathbb{R}^{d}$ who contains comprehensive information from all query qualifiers. Apart from $\mathbf{h}^\text{que}_\text{Qual}$, we also devise a qualifier matcher that further exploits additional supporting information from the qualifiers of other observed facts related to query subject $s$ in $\mathcal{G}_\text{obs}$. Qualifier matcher finds all the HTKG facts in $\mathcal{G}_\text{obs}$ where each of them takes $s$ as the subject of its primary quadruple\footnote{We only consider subject-related qualifiers because we can only observe the subject entity in each LP query and we aim to find the additional qualifiers most related to the query.}. It then collects all their qualifiers $\{(\Bar{r}_{q_l}, \Bar{e}_{q_l})\}_{l=1}^{n_\text{all}}$ and computes a global qualifier feature
\begin{equation}
\small
\resizebox{0.9\columnwidth}{!}{%
$
\begin{aligned}
    &\eta_l =  \frac{\text{exp}((\mathbf{W}_3(\mathbf{h}_{\Bar{r}_{q_l}}\| \mathbf{h}_{\Bar{e}_{q_l}}))^\top (
    \mathbf{W}_4(\mathbf{h}_{(s,t)}\| \mathbf{h}_r)))}{\sum_{m=1}^{n_\text{all}} \text{exp}((\mathbf{W}_3(\mathbf{h}_{\Bar{r}_{q_m}}\| \mathbf{h}_{\Bar{e}_{q_m}}))^\top (\mathbf{W}_4(\mathbf{h}_{(s,t)}\| \mathbf{h}_r)))},\\
    &\mathbf{h}_\text{Qual}^\text{glo} =  \sum_{q_l} \eta_l \mathbf{W}_3(\mathbf{h}_{\Bar{r}_{q_l}}\| \mathbf{h}_{\Bar{e}_{q_l}}),
\end{aligned}
$}
\end{equation}
where $n_\text{all}$ denotes the number of $s$-related qualifiers and $\mathbf{W}_3, \mathbf{W}_4 \in \mathbb{R}^{d\times 2d}$ are weight matrices. $\mathbf{h}_{(s,t)} = f_t(\mathbf{h}_s \| \mathbf{h}_t)$. $\eta_l$ is the attention score of the $l^\text{th}$ subject-related qualifier indicating its contribution to the LP query. Given $\mathbf{h}_\text{Qual}^\text{que}$ and $\mathbf{h}_\text{Qual}^\text{glo}$ ($\mathbf{h}_\text{Qual}^\text{glo} \in \mathbb{R}^d$), QMD uses another query-wise Transformer to compute a query feature. We concatenate the representation of another separate [CLS] token with $\mathbf{h}_{(s,t)} \| \mathbf{h}_r \| \mathbf{h}^\text{que}_\text{Qual} \| \mathbf{h}^\text{glo}_\text{Qual}$ and input it into the query-wise Transformer. The output representation of this separate [CLS] token corresponds to $\mathbf{h}^\text{que} \in \mathbb{R}^d$. $\mathbf{h}^\text{que}$ is used by QMD to compute a score for each candidate entity $e_c \in \mathcal{E}$
\begin{equation}
\label{eq: score}
\small
\resizebox{0.6\columnwidth}{!}{%
$
    \lambda(e_c) = {(\mathbf{h}^\text{que} * \mathbf{h}_t)}^\top \mathbf{W}_5 \mathbf{h}_{e_c}.
$}
\end{equation}
$\mathbf{W}_5 \in \mathbb{R}^{d\times d}$ is a score matrix. 
HypeTKG takes the candidate entity with the highest score as the predicted answer.
\subsection{Time-Invariant Knowledge Modeling}
\label{sec: static model}
Previous sections
discuss how HypeTKG performs HTKG LP without using TI knowledge. In this section, we discuss how we adapt HypeTKG to TI knowledge by developing a model variant HypeTKG$^\psi$. We first introduce another gated structure in QATGE to incorporate TI knowledge in the encoding process. We change Eq. \ref{eq: QATGE} to
\begin{equation}
\small
\resizebox{0.9\columnwidth}{!}{%
$
\begin{aligned}
&\mathbf{h}_e^\text{temp} = \frac{1}{|\mathcal{N}_e|}\sum_{\zeta \in \mathcal{N}_e} \mathbf{W}_2 \phi\left(\mathbf{h}_{(e', t')}, \left(\gamma \mathbf{h}^\zeta_\text{Qual} + (1-\gamma)\mathbf{h}_{r'}\right)\right),\\
&\mathbf{h}^\psi_e = \frac{1}{|\mathcal{N}_e^\psi|} \sum_{\zeta^\psi \in \mathcal{N}_e^\psi} \mathbf{W}^\psi \phi(\mathbf{h}_{e''}, \mathbf{h}_{r''}),\\
&\mathbf{h}_e = (1-\beta) \mathbf{h}_e^\text{temp} + \beta \mathbf{h}^\psi_e.
\end{aligned}
$}
\end{equation}
$\beta$ is a trainable parameter controlling the magnitude of TI information. $\mathcal{N}_e^\psi = \{\zeta^\psi\} = \{(e'', r'')|(e'', r'', e) \in \mathcal{G}_\text{TI}\}$ denotes $e$'s TI neighbors derived from additional TI facts. $\mathbf{h}_e^\text{temp}$ and $\mathbf{h}^\psi_e$ contain the encoded temporal and TI information, respectively. In QMD, we incorporate TI knowledge when we compute the query feature $\mathbf{h}^\text{que}$. Same as how we model query qualifiers, we use a TI-wise Transformer to model $s$'s TI neighbors and output a TI feature $\mathbf{h}^s_\text{TI}$. We expand the input length of the query-wise Transformer and input $\mathbf{h}_{(s,t)} \| \mathbf{h}_r \| \mathbf{h}^\text{que}_\text{Qual} \| \mathbf{h}^\text{glo}_\text{Qual} \| \mathbf{h}^s_\text{TI}$ for computing $\mathbf{h}^\text{que}$. Note that we do not model TI neighbors of all $|\mathcal{E}|$ candidate entities in QMD because (1) this will incur excessive computational cost and (2) this part of information has been learned in QATGE.
\subsection{Parameter Learning}
We minimize a binary cross-entropy (BCE) loss for learning model parameters. We take every fact in $\mathcal{G}_\text{obs}$
as a query fact $\delta$ and switch its object entity $o$ to every other entity $e \in (\mathcal{E}\setminus\{o\})$ to create $|\mathcal{E}|-1$ negative facts $\{\delta^-\}$. Our loss is defined as
\begin{equation}
\small
\resizebox{0.85\columnwidth}{!}{%
$
    \mathcal{L} = \frac{1}{|\mathcal{G}_\text{obs}| \times |\mathcal{E}|} \sum_{\delta \in \mathcal{G}_\text{obs}} (l_{\delta} + \sum_{\delta^-} l_{\delta^-}).
$}
\end{equation}
$l_{\delta} =  -y_{\delta} \log (\lambda(\delta)) - (1-y_{\delta})\log (1 - \lambda(\delta))$, $l_{\delta^-} = -y_{\delta^-} \log (\lambda(\delta^-))-(1-y_{\delta^-}) \log (1-\lambda(\delta^-) )$ denote the BCE of $\delta$ and $\delta^-$, respectively. $y_{\delta} = 1$ and $y_{\delta^-} = 0$ because we want to simultaneously maximize $\lambda(\delta)$ and minimize $\lambda(\delta^-)$. $|\mathcal{G}_\text{obs}|$ is the number of HTKG facts in $\mathcal{G}_\text{obs}$.


\section{Experiments}
We do HTKG LP over Wiki-hy and YAGO-hy. We report HTKG LP results in Sec. \ref{sec: main results}. We study whether additional TI knowledge helps HTKG LP in Sec. \ref{sec: static result}. We do ablation studies and study the impact of the ratio of utilized qualifiers in Sec. \ref{sec: further analysis}. Finally, we present several case studies to show the effectiveness of leveraging TI knowledge and qualifier matcher for temporal fact reasoning over HTKGs in Sec. \ref{sec: case}. We provide complexity analysis of our model in App. \ref{app: complexity}. We also study the impact of qualifier-augmented fact proportion and present it in App. \ref{app: proportion}.
\subsection{Experimental Setting}
We use two evaluation metrics, i.e., mean reciprocal rank (MRR) and Hits@1/3/10. 
We follow the filtering setting used in previous HKG reasoning works \cite{DBLP:conf/emnlp/GalkinTMUL20}. See App. \ref{app: metric} for detailed explanations of evaluation metrics.
We consider two types of baselines: (1) Traditional TKG interpolation methods\footnote{TKG extrapolation methods are not considered since we only study interpolated LP over HTKGs. Extrapolation methods are constrained to only use the graph information before each LP query, making them suboptimal for interpolation.}, i.e., DE-SimplE \cite{DBLP:conf/aaai/GoelKBP20}, TeRo \cite{DBLP:conf/coling/XuNAYL20}, T-GAP \cite{DBLP:conf/kdd/JungJK21}, BoxTE \cite{DBLP:conf/aaai/MessnerAC22}, TARGCN \cite{ding2022a}, TeAST \cite{DBLP:conf/acl/LiSG23} and HGE \cite{DBLP:conf/aaai/PanNLS24}.
Since these methods have no way to model qualifiers, we neglect the qualifiers during implementation. (2) HKG reasoning methods, i.e., NaLP-Fix \cite{DBLP:conf/www/RossoYC20}, HINGE \cite{DBLP:conf/www/RossoYC20}, HypE \cite{DBLP:conf/ijcai/FatemiTV020}, StarE \cite{DBLP:conf/emnlp/GalkinTMUL20}, GRAN \cite{DBLP:conf/acl/WangWLZ21}, HyconvE \cite{DBLP:conf/www/WangWLC023}, ShrinkE \cite{DBLP:conf/acl/XiongNPS23} and HyNT \cite{DBLP:conf/kdd/ChungLW23}. These methods cannot model temporal information in HTKGs. We make them neglect the timestamps during implementation.
See App. \ref{app: implementation detail} for HypeTKG and baseline implementation details. 
Note that NE-Net \cite{DBLP:conf/emnlp/HouJL0GZGC23} still has no existing software and data, so we are unable to directly compare it with HypeTKG here.

\subsection{Comparative Study}
\label{sec: main results}
We report the HTKG LP results of all methods in Table \ref{tab: tid LP results}. We observe that HypeTKG outperforms all baselines and achieves state-of-the-art. We believe this is because (1) traditional TKG reasoning methods lose a large amount of semantic information by failing to model qualifiers (2) and previous HKG reasoning baselines cannot distinguish from different timestamps, which is key to temporal fact reasoning. We also observe that HypeTKG$^\psi$ achieves even better results than the original model. We will have a more detailed discussion in Sec. \ref{sec: static result}.
\begin{table}[htbp]
    \centering
    \resizebox{\columnwidth}{!}{
    \large\begin{tabular}{@{}lcccccccc@{}}
\toprule
        \textbf{Datasets} & \multicolumn{4}{c}{\textbf{WiKi-hy}} &  \multicolumn{4}{c}{\textbf{YAGO-hy}} \\
\cmidrule(lr){2-5} \cmidrule(lr){6-9}
        \textbf{Model} & MRR & H@1 & H@3 & H@10 & MRR & H@1 & H@3  & H@10 \\
\midrule 
        DE-SimplE 
       & 0.351	& 0.218 & 0.405	& 0.640
       & 0.684	& 0.625	& 0.715 & 0.807
       \\
        TeRo
        & 0.572	& 0.473	& 0.640	& 0.727
        & 0.760	& 0.720	& 0.782	& 0.822
        \\
        T-GAP 
        & 0.588	& 0.486	& 0.651	& 0.726
        & 0.773	& 0.736	& 0.800	& 0.835
        \\
        BoxTE 
        & 0.449	& 0.348	& 0.512	& 0.646
        & 0.685	& 0.642	& 0.725	& 0.767 
        \\
        TARGCN 
        & 0.589	& 0.498	& 0.652	& 0.733
        & 0.769	& 0.742	& 0.772	& 0.817
        \\
        TeAST 
        & 0.601	& 0.507	& 0.669	& 0.761
        & 0.794	& 0.763	& 0.817	& 0.844
        \\
        HGE 
        & 0.602	& 0.507	& 0.666	& 0.765
        & 0.790	& 0.760	& 0.814	& 0.837
        \\
\midrule
       NaLP-Fix
       & 0.507	& 0.460	& 0.569	& 0.681
       & 0.730	& 0.709	& 0.751 & 0.813
       \\
        HINGE 
       & 0.543	& 0.497	& 0.585	& 0.694
       & 0.758	& 0.730	& 0.762 & 0.819
       \\
       HypE
       & 0.624	& 0.604	& 0.631	& 0.658
       & 0.800	& 0.785	& 0.799 & 0.830
       \\
       StarE 
       & 0.565	& 0.491	& 0.599	& 0.703
       & 0.765	& 0.737	& 0.776	& 0.820
       \\
       GRAN & 0.661 & 0.610 & 0.679 & 0.750 
       & 0.808 & 0.789 & 0.817 & 0.842 
        \\
       HyconvE 
       & 0.641	& 0.600	& 0.656	& 0.729
       & 0.771	& 0.754	& 0.782	& 0.811 
       \\
        ShrinkE & 0.669 & 0.593 & 0.703 & \textbf{0.789} 
       & 0.808 & 0.782 & 0.824 & 0.852 
       \\
       HyNT
       & 0.537	& 0.444	& 0.587	& 0.723
       & 0.763	& 0.724	& 0.787	& 0.836
       \\
\midrule
        HypeTKG
        & \textbf{0.687} & \textbf{0.633} & \textbf{0.710} & \textbf{0.789}
        & \textbf{0.832} & \textbf{0.817} & \textbf{0.838} & \textbf{0.857}
        \\
        \midrule
        HypeTKG$^\psi$
        & 0.693 & 0.642 & 0.715 & 0.792
        & 0.842  & 0.821& 0.839 & 0.858
        \\
\bottomrule
    \end{tabular}
    }
    \caption{HTKG LP results.
    The best results without using TI facts are marked in bold. H@1/H@3/H@10 means Hits@1/Hits@3/Hits@10.}
\label{tab: tid LP results}
\end{table}
\subsection{Do TI Relational Knowledge Help HTKG Reasoning?}
\label{sec: static result}
We let HypeTKG and all baselines use the additional TI facts and report their temporal fact LP performance on Wiki-hy and YAGO-hy in Table \ref{tab: static analysis}. For the HKG approaches, we directly include these facts into our datasets. For traditional TKG reasoning approaches, we create a number of temporal facts for each TI fact along the whole timeline and include these temporal facts into the datasets. For example, let $t_\text{min}$/$t_\text{max}$ denotes the minimum/maximum timestamp of an HTKG. We transform a TI fact $(s,r,o)$ to $\{(s,r,o,t_\text{min}), ..., (s,r,o,t_\text{max})\}$. 
Surprisingly, we observe that while HypeTKG constantly benefit from the additional TI relational knowledge, other baselines cannot. 
We attribute this to the following reasons: (1) TI facts introduce distributional shift. Baseline methods learn TI and temporal knowledge without distinguishing their difference, making them less focused on the temporal facts.
(2) HypeTKG employs its gate-structured graph encoder that adaptively controls the amount of information from the TI facts. HypeTKG's decoder also uses Transformer to distinguish the importance of different TI facts. These two steps help HypeTKG to exploit the TI knowledge that is most beneficial in LP and discard the redundant information. We further study whether TI knowledge can improve reasoning on quadruple-based TKGs in App. \ref{app: ti for tkg}.
\begin{table}[htbp]
    \centering
    \resizebox{0.85\columnwidth}{!}{
    \large\begin{tabular}{@{}lcccccc@{}}
\toprule
        \textbf{Datasets} & \multicolumn{3}{c}{\textbf{WiKi-hy}} &  \multicolumn{3}{c}{\textbf{YAGO-hy}} \\
\cmidrule(lr){2-4} \cmidrule(lr){5-7}
        \textbf{Model} & w.o. TI & w. TI & $\Delta\uparrow$ & w.o. TI & w. TI & $\Delta\uparrow$ \\
\midrule 
        DE-SimplE
        & 0.351	& 0.326	& \textcolor{Red}{-0.025}
        & 0.684	& 0.643	& \textcolor{Red}{-0.041}	
        \\
        TeRo
        & 0.572	& 0.553	& \textcolor{Red}{-0.019}
        & 0.760	& 0.742	& \textcolor{Red}{-0.018}	
        \\
        T-GAP 
        & 0.588	& 0.568	& \textcolor{Red}{-0.020}
        & 0.773	& 0.761	& \textcolor{Red}{-0.012}
        \\
        BoxTE 
        & 0.449	& 0.409	& \textcolor{Red}{-0.040}
        & 0.685	& 0.670	& \textcolor{Red}{-0.015}
        \\
        TARGCN 
        & 0.589	& 0.588	& \textcolor{Red}{-0.001}
        & 0.769	& 0.769	& \textcolor{Red}{0.000}
        \\
        TeAST 
        & 0.601	& 0.581	& \textcolor{Red}{-0.020}
        & 0.794	& 0.779	& \textcolor{Red}{-0.015}
        \\
        HGE 
        & 0.602	& 0.592	& \textcolor{Red}{-0.010}
        & 0.790	& 0.780	& \textcolor{Red}{-0.010}
        \\
\midrule
        NaLP-Fix
       & 0.507	& 0.504	& \textcolor{Red}{-0.003}
       & 0.730	& 0.728	& \textcolor{Red}{-0.002}	
        \\
        HINGE 
       & 0.543	& 0.535	& \textcolor{Red}{-0.008}
       & 0.758	& 0.754	& \textcolor{Red}{-0.004}
       \\
       HypE 
       & 0.624	& 0.623 & \textcolor{Red}{-0.001}
       & 0.800	& 0.798	& \textcolor{Red}{-0.002}
       \\
       StarE 
       & 0.565	& 0.547	& \textcolor{Red}{-0.018}
       & 0.765	& 0.758	& \textcolor{Red}{-0.007}
       \\
        GRAN & 0.661 & 0.667 & \textcolor{Green}{\textbf{+0.006}}
       & 0.808 & 0.794 & \textcolor{Red}{-0.014} 
        \\
       HyconvE 
       & 0.641	& 0.630	& \textcolor{Red}{-0.011}
       & 0.771	& 0.767	& \textcolor{Red}{-0.004}	 
       \\
        ShrinkE & 0.669 & 0.655 & \textcolor{Red}{-0.014}
       & 0.808 & 0.806 & \textcolor{Red}{-0.002}  
       \\
       HyNT 
       & 0.537	& 0.536	& \textcolor{Red}{-0.001}
       & 0.763	& 0.765	& \textcolor{Green}{+0.002}	 
       \\
\midrule
        HypeTKG
        & \textbf{0.687} & \textbf{0.693}  & \textcolor{Green}{\textbf{+0.006}}
        & \textbf{0.832} & \textbf{0.842}  & \textcolor{Green}{\textbf{+0.010}}  
        \\
\bottomrule
    \end{tabular}
    }
    \caption{MRR for all methods with (w. TI) and without (w.o. TI) TI facts. $\Delta\uparrow$ denotes the absolute improvement. Note that HypeTKG w. TI equals HypeTKG$^\psi$.}
\label{tab: static analysis}
\end{table}
\subsection{Further Analysis}
\label{sec: further analysis}
\paragraph{Ablation Study}
\label{sec: ablation}
We conduct ablation studies to demonstrate the importance of different model components of HypeTKG. In study A (Variant A), we neglect the qualifiers in all HTKG facts and do not include any qualifier learning component. In study B (Variant B), we remove qualifier attention in QATGE. In study C (Variant C), we remove the qualifier matcher in QMD. In study D (Variant D), we exclude time modeling modules and neglect timestamps in primary quadruples. From Table \ref{tab: proportion}, we observe that learning qualifiers is essential in reasoning HTKGs. Both qualifier attention in QATGE and qualifier matcher contribute to qualifier modeling. We also find that modeling temporal information is essential for temporal fact reasoning.
\begin{table}[htbp]
    \centering
    \resizebox{\columnwidth}{!}{
    \large\begin{tabular}{@{}l|ccc|cccccc@{}}
\toprule
        & \multicolumn{3}{c|}{\textbf{Setting}} & \multicolumn{3}{c}{\textbf{Wiki-hy}} & \multicolumn{3}{c}{\textbf{YAGO-hy}}\\
\cmidrule(lr){2-4} \cmidrule(lr){5-7} \cmidrule(lr){8-10}
        \textbf{Model} & Time & Q Att & Q Match & MRR & H@1 & H@10 & MRR & H@1 & H@10 \\
\midrule 
        Variant A  & \ding{51} & \ding{55}& \ding{55}& 0.642 & 0.569 & 0.775 
        & 0.795 & 0.770  & 0.841

         \\
        Variant B  & \ding{51} & \ding{55} & \ding{51}& 0.671 & 0.616 & 0.777 
        & 0.826 & 0.811  & 0.856 

         \\
        Variant C & \ding{51} & \ding{51} & \ding{55} & 0.671 & 0.615 & 0.777 
        & 0.803 & 0.781  & 0.842

         \\
         Variant D & \ding{55} & \ding{51} & \ding{51} 
         & 0.652 & 0.597  & 0.751
         & 0.792  & 0.769  & 0.835

         \\
        HypeTKG & \ding{51} & \ding{51}& \ding{51}&  \textbf{0.687} & \textbf{0.633} & \textbf{0.789} 
        &  \textbf{0.832} & \textbf{0.817}  & \textbf{0.857}

        \\

\bottomrule
    \end{tabular}
    }
    \caption{Ablation studies. Q means qualifier.
    }
    \label{tab: proportion}
\end{table}
\begin{figure}[htbp]
    \centering
       \includegraphics[width=0.8\columnwidth]{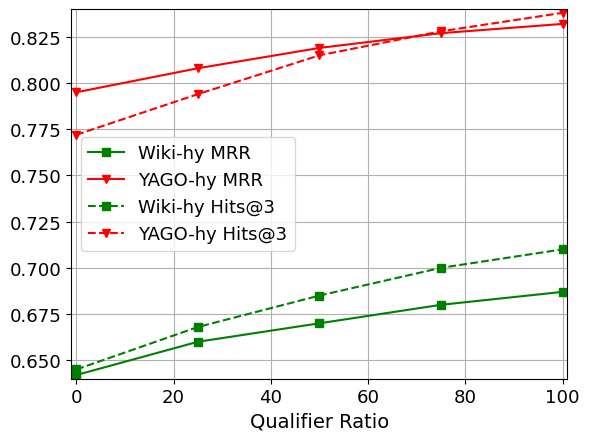}
  \caption{HypeTKG performance with a varying ratio of used qualifiers.}
  \label{fig: qualifier ratio} 
\end{figure}
\begin{table*}[htbp]
\small
    \centering
    \resizebox{\textwidth}{!}{
\begin{tabular}{c| c| c| c| c} 
\toprule
\textbf{Case}&\textbf{Query}&\textbf{Answer}&\textbf{Subject-Related Qualifiers}&\textbf{Attention Score}
\\ 
\midrule
\multirow{3}{*}{A1}&\multirow{3}{*}{$((\textit{Andrey Kolmogorov}, \textit{award received}, ?, 1941), \emptyset)$}  &	\multirow{3}{*}{$\textit{USSR State Prize}$} & $(\textit{country of citizenship}, \textit{Soviet Union})$ & $9.39e^{-1}$  \\

&&& $(\textit{field of work}, \textit{mathematics})$ & $6.09e^{-2}$\\
&&& $(\textit{country}, \textit{Soviet Union})$ & $2.61e^{-10}$\\
\midrule
\multirow{3}{*}{A2}&\multirow{2}{*}{$((\textit{Andrey Kolmogorov}, \textit{place of death}, ?, 1987),$}  &	\multirow{3}{*}{$\textit{Moscow}$} & $(\textit{country of citizenship}, \textit{Soviet Union})$ & $0.99$  \\

&&& $(\textit{field of work}, \textit{mathematics})$ & $1.64e^{-21}$\\
&$\{(\textit{country}, \textit{Soviet Union})\})$&& $(\textit{country}, \textit{Soviet Union})$ & $5.00e^{-22}$\\

\bottomrule
\end{tabular}
}
\caption{Case study A: cases for studying qualifier matcher.}
\label{tab: case}
\end{table*}

\begin{table*}[htbp]
\small
    \centering
    \resizebox{\textwidth}{!}{
\begin{tabular}{c| c| c| c| c} 
\toprule
\textbf{Case}&\textbf{Query}&\textbf{Prediction w. TI}&\textbf{Prediction w.o. TI} & \textbf{Related TI Facts}
\\ 
\midrule
\multirow{3}{*}{B1}&\multirow{3}{*}{$((\textit{Pisa}, \textit{country}, ?, 1860), \emptyset)$}  &	\multirow{3}{*}{$\textit{Kingdom of Sardinia}$} & \multirow{3}{*}{$\textit{Kingdom of Prussia}$}  & $(\textit{Pisa}, \textit{official language}, \textit{Italian})$\\
&&&& $(\textit{Kingdom of Sardinia}, \textit{official language}, \textit{Italian})$\\
&&&& $(\textit{Kingdom of Prussia}, \textit{official language}, \textit{German})$\\
\midrule
\multirow{2}{*}{B2}&\multirow{2}{*}{$((\textit{AK}, \textit{place of birth}, ?, 1903), \{(\textit{country}, \textit{Russian Empire})\})$}  &	\multirow{2}{*}{$\textit{Tbilisi}$} & \multirow{2}{*}{$\textit{Moscow}$} & $(\textit{AK}, \textit{native language}, \textit{Georgian})$\\
&&&& $(\textit{Tbilisi}, \textit{official language}, \textit{Georgian})$\\
\bottomrule
\end{tabular}
}
\caption{Case study B: cases for studying the effectiveness of TI relational knowledge. Prediction w./w.o. TI means the prediction result with/without using time-invariant facts. \textit{AK} is the abbreviation of the entity \textit{Aram Khachaturian}.}
\label{tab: case2}
\end{table*}
\paragraph{Impact of the Ratio of Utilized Qualifiers}
\label{sec: vary ratio}
To further investigate the importance of learning qualifiers for reasoning hyper-relational temporal facts, we report HypeTKG's performance on Wiki-hy/YAGO-hy by using a varying ratio of utilized qualifiers. We implement HypeTKG on all Wiki-hy/YAGO-hy facts but randomly sample 0\%/25\%/50\%/75\%/100\% of all the existing qualifiers during training and evaluation. From Fig. \ref{fig: qualifier ratio}, we observe that HypeTKG achieves better results as the ratio increases, showing a positive correlation between its performance and the number of utilized qualifiers. This indicates that modeling qualifiers is beneficial for LP over temporal facts.

\subsection{Case Studies}
\paragraph{A: Effectiveness of Qualifier Matcher}
\label{sec: case}
We do case studies to show how our qualifier matcher improves HTKG reasoning (Table \ref{tab: case}). HypeTKG ranks the ground truth missing entities in these cases as top 1. As discussed in Sec. \ref{sec: decoder}, the qualifier matcher interprets the contribution of all the existing qualifiers related to the subject entity of the LP query with attention scores $\eta_l$. 
In Case A1, no qualifier is provided in the query for prediction.
We find that qualifier matcher assigns a great attention score to the qualifier $(\textit{country of citizenship}, \textit{Soviet Union})$ from another fact.
It can be taken as a hint to predict the ground truth missing entity \textit{USSR State Prize}.
This implies that to better reason the facts without qualifiers, our qualifier matcher can find the clues from other hyper-relational facts. 
In Case A2, we find that the qualifier matcher focuses more on the qualifiers from other facts but not from the query. Note that the query qualifiers have been modeled with a query-specific qualifier feature $\mathbf{h}^\text{que}_\text{Qual}$ before computing the global qualifier feature. This indicates that our qualifier matcher can maximally extract information from the extra qualifiers rather than only focusing on the query qualifiers, enabling efficient information fusion. See App. \ref{app: case} for more case study details and one more case (A3) discussion.
\paragraph{B: Effectiveness of TI Knowledge}
\label{sec: case2}
We demonstrate how TI relational knowledge enhances HTKG reasoning with two cases (Table \ref{tab: case2}). In both cases, HypeTKG achieves optimal prediction by leveraging TI knowledge, and makes mistakes without it. 
In B1, HypeTKG predicts the false answer $\textit{Kingdom of Prussia}$ without the support of TI facts. However, after considering them, HypeTKG manages to make accurate prediction because $\textit{Pisa}$ should share the same official language with the country that contains it. 
In B2, since both \textit{Tbilisi} and \textit{Moscow} belonged to \textit{Russian Empire} in 1903, it is hard for HypeTKG to distinguish them during prediction without any further information. However, by knowing that \textit{Aram Khachaturian}'s native language is same as the official language of \textit{Tbilisi}, i.e., Georgian, HypeTKG can exclude the influence of \textit{Moscow} because people speak Russian there.

\section{Conclusion}
In this work, we propose a new data structure named HTKG for studying temporal fact reasoning over HKGs. 
To reason HTKGs, we design a model HypeTKG that is able to simultaneously deal with temporal information and qualifiers. 
We benchmark HypeTKG and various previous HKG/TKG reasoning methods on two newly-constructed datasets, i.e., Wiki-hy and YAGO-hy.
We show that HypeTKG achieves superior performance on HTKG LP. Besides, we mine the TI relational knowledge from Wikidata KB and study whether it can benefit models on hyper-relational temporal fact reasoning. We find that temporal fact reasoning on HTKGs can be enhanced by carefully balancing the information between temporal and TI knowledge.  
\section{Limitations}
One limitation of our work is that we have not explored qualifier prediction, i.e., predicting the missing elements in the qualifiers. We also have not considered another challenge in temporal fact reasoning, i.e., time prediction. We think our work can be the base of future studies on these two topics. Also, as we have only studied interpolated link prediction on HTKGs, developing HTKG extrapolation methods would also be an important direction in the future. Besides, given the growing interest in inductive learning on traditional TKGs (e.g., \cite{ding2022few,DBLP:conf/pkdd/DingWLMT23,DBLP:conf/ijcnn/DingHWMHT23}, we believe it is equally important to explore inductive learning on HTKGs, which remains unaddressed in this work.

\section*{Acknowledgement}
This work has been funded by the Munich Center for Machine Learning and supported by the Federal Ministry of Education and Research and the State of Bavaria. Jingcheng Wu has been funded by the Deutsche Forschungsgemeinschaft (DFG, German Research Foundation) - SFB 1574 - Project number 471687386. The authors thank the International Max Planck Research School for Intelligent Systems (IMPRS-IS) for supporting Bo Xiong. 
This research has also been partially funded by Deutsche Forschungsgemeinschaft (DFG, German Research Foundation) under Germany’s Excellence Strategy - EXC 2075 - 390740016, the Stuttgart Center for Simulation Science (SimTech), the European Union’s Horizon 2020 research and the the Bundesministerium für Wirtschaft und Energie (BMWi), grant aggrement No. 01MK20008F.

\bibliography{custom}

\appendix
\section{Additional HTKG example}
\label{app: more htkg example}
\begin{figure}[htbp]
    \centering
    \includegraphics[width=0.9\columnwidth]{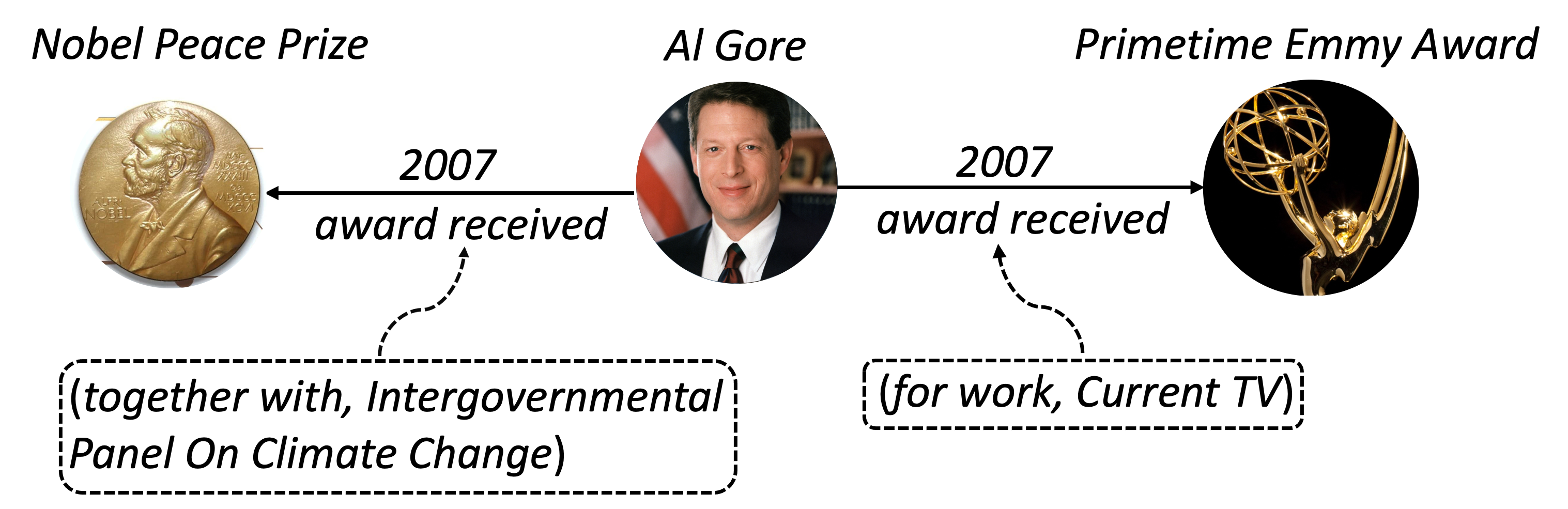}
    \caption{Additional example of HTKG fact.}
    \label{fig: hyper example2}
\end{figure}

We provide an additional HTKG fact highlighting the diversity of qualifiers within distinct temporal facts happening at the same timestamp (Fig. \ref{fig: hyper example2}). The two awards \textit{Al Gore} received in \textit{2007} can be differentiated considering the coupled qualifiers.
\section{YAGO-hy Construction Details}
\label{app: yago mapping}
We provide the relation mapping from YAGO1830 to Wikidata in Table \ref{tab: yago relation mapping}. During matching, we carefully examine YAGO1830 facts and find that \textit{playsFor} represents a person playing for a sports team, and \textit{isAffiliatedTo} represents a person's political affiliation. Therefore, we map \textit{playsFor} to \textit{member of sports team} and \textit{isAffiliatedTo} to \textit{member of political party}. Besides, YAGO1830 is originally a TKG extrapolation dataset, we redistribute its facts and change it into an interpolation dataset before qualifier searching. We ensure that the proportions of the number of facts in train/valid/test sets of YAGO-hy conform to the corresponding sets in YAGO1830.
\begin{table}[htbp] 
    \begin{center}
      \resizebox{\columnwidth}{!}{
    \begin{tabular}{c|cc} 
      \toprule 
     \multicolumn{1}{c|}{\textbf{YAGO Relation}} & \multicolumn{1}{c}{\textbf{Wikidata Relation}} & \multicolumn{1}{c}{\textbf{Wikidata Relation ID}}  \\
      \midrule 
      wasBornIn & place of birth    &  P19\\
      diedIn &  place of death   &  P20\\
      worksAt &  employer &  P108\\
      playsFor & member of sports team & P54\\
      hasWonPrize & award received & P166\\
      isMarriedTo & spouse & P26\\
      owns & owned by$^{-1}$ & P127\\
      graduatedFrom & educated at & P69\\
      isAffiliatedTo & member of political party & P102\\
      created & notable work & P800\\
       
      \bottomrule 
    \end{tabular} }
    \end{center}
    \caption{Relation type mapping from YAGO1830 to Wikidata. \textit{owned by}$^{-1}$ denotes the inverse relation of \textit{owns}}
    \label{tab: yago relation mapping}
\end{table}
\section{Why Not Construct ICEWS-Based HTKGs?}
\label{app: why not icews}
Integrated Crisis Early Warning System (ICEWS)  \cite{DVN/28075_2015} is another popular KB for constructing quadruple-based TKGs. Hou et al. \cite{DBLP:conf/emnlp/HouJL0GZGC23} use ICEWS to construct an N-TKG, i.e., NICE. We do not use ICEWS to construct HTKGs due to the following reasons. Different from Wikidata, every fact in ICEWS has no additional statements that can serve as qualifiers. To solve this problem, Hou et al. design rule templates on ICEWS relations and decompose the relation of each ICEWS quadruple-based fact into several parts. For example, an ICEWS-based fact $(\textit{Iran}, \textit{express intent to provide humanitarian aid}, \\\textit{Yemen}, t)$ will be transformed into: 
\begin{equation*}
    \begin{aligned}
        &(\textit{express intent to cooperate}, \\
        &\textit{volunteer}:\textit{Iran}, \\ 
        &\textit{cooperation target}:\textit{Yemen}, \\
        &\textit{cooperation content}:\textit{provide humanitarian aid}, \\
        &t).
    \end{aligned}
\end{equation*}
N-TKG assumes that this transformation brings auxiliary information into fact quadruples, however, we think the amount of additional information is highly limited. This is because the transformation from an ICEWS-based fact quadruple into an N-TKG fact does not consider any additional information source other than the original quadruple. In other words, the amount of information stored in an ICEWS-based fact quadruple is nearly the same as the amount carried by its $n$-tuple form. As discussed in previous works about HKGs, qualifiers are introduced to better restrict the fact validity and also increase the data expressiveness. Due to the lack of additional linked statements in ICEWS, it is not easy to construct meaningful HTKGs based on this KB.

\section{Complexity Analysis} 
\label{app: complexity}
The time complexity of HypeTKG is the same as most of previous GNN-based TKG approaches, which is $O(|\mathcal{T}||\mathcal{E}| + |\mathcal{T}||\mathcal{R}|)$, where $\mathcal{T}$, $\mathcal{E}$, and $\mathcal{R}$ are the number of timestamps, entities, and relations, respectively. Similarly, the memory complexity is $O(|\mathcal{E}|d + |\mathcal{R}|d)$. The qualifier modeling modules, though requires additional computation, does not increase the time and memory complexity as qualifiers are also composed by entities and relations. As for HypeTKG$^\psi$, since it considers time-invariant knowledge that introduces additional entities and relations, the time complexity becomes $O(|\mathcal{T}|(|\mathcal{E}| + |\mathcal{E}_\text{TI}|) + |\mathcal{T}|(|\mathcal{R}| + |\mathcal{R}_\text{TI}|))$ and the memory complexity is $O((|\mathcal{E}| + |\mathcal{E}_\text{TI}|)d + (|\mathcal{R}| + |\mathcal{R}_\text{TI}|)d)$. $|\mathcal{E}_\text{TI}|$ and $|\mathcal{R}_\text{TI}|$ are the numbers of introduced new entities and relations in time-invariant facts, respectively.

\section{Evaluation Metrics Details}
\label{app: metric}
MRR computes the mean of the reciprocal ranks for all test queries: $\frac{1}{2N_{\text{test}}}\sum_{\text{que}} \frac{1}{\theta_\text{que}}$, where $\theta_\text{que}$ denotes the rank of the ground truth missing entity in the test query $\text{que}$. Note that for each fact in the test set, we derive two LP queries for both subject and object entity prediction, and therefore, the total number of test queries is $2N_{\text{test}}$. Hits@1/3/10 denotes the proportion of the test queries where ground truth entities are ranked as top 1/3/10.
\section{Complex Vector Mapping Details}
\label{app: complex vector}
$\mathbf{h}^\mathbb{C}_{e'_{q_i}}  \in \mathbb{C}^{\frac{d}{2}}$ and $\mathbf{h}^\mathbb{C}_{r'_{q_i}} \in \mathbb{C}^{\frac{d}{2}}$ are the complex vectors mapped from $\mathbf{h}_{e'_{q_i}}$ and $\mathbf{h}_{r'_{q_i}}$. The real part of $\mathbf{h}^\mathbb{C}_{e'_{q_i}}$ is the first half of $\mathbf{h}_{e'_{q_i}}$ and the imaginary part is the second half, e.g., if $\mathbf{h}_{e'_{q_i}} = [6,3]^\top \in \mathbb{R}^2$, then $\mathbf{h}^\mathbb{C}_{e'_{q_i}} = [6+3\sqrt{-1}]^\top \in \mathbb{C}^1$. $\mathbf{h}_{r'_{q_i}}^{\mathbb{C}}[j] = \cos(\mathbf{h}_{r'_{q_i}}[j]) + \sqrt{-1}\sin(\mathbf{h}_{r'_{q_i}}[\frac{d}{2}+j])$, where $\mathbf{h}_{r'_{q_i}}^{\mathbb{C}}[j]$ and $\mathbf{h}_{r'_{q_i}}[\frac{d}{2} + j]$ denote the $j^\text{th}$ and $(\frac{d}{2} + j)^\text{th}$ element of $\mathbf{h}_{r'_{q_i}}^{\mathbb{C}}$ and $\mathbf{h}_{r'_{q_i}}$, respectively.

\section{Implementation Details}
\label{app: implementation detail}
We implement all the experiments of HypeTKG and baselines with PyTorch \cite{DBLP:conf/nips/PaszkeGMLBCKLGA19} on an NVIDIA A40 with 48GB memory and a 2.6GHZ AMD EPYC 7513 32-Core Processor. For HypeTKG, we set the batch size to 256 and use the Adam optimizer with an initial learning rate of 0.0001. We search hyperparameters following Table \ref{tab: hyperparameter search}. For each dataset, we do 108 trials to try different hyperparameter settings. We run 100 epochs for each trial and compare their validation results. We choose the setting leading to the best validation result and take it as the best hyperparameter setting. The best hyperparameter setting is also stated in Table \ref{tab: hyperparameter search}.
Every result reported is the average result of five runs with different random seeds. The error bars are relatively small and are omitted. We report the total training time of our model until it reaches maximum performance in Table \ref{tab: train time}. We also specify the GPU memory usage (Table \ref{tab: memory}) and number of parameters (Table \ref{tab: num_param}).
\begin{table}[t] 
\centering

    \resizebox{0.95\columnwidth}{!}{
    \begin{tabular}{cc} 
      \toprule 
     \textbf{Hyperparameter} & \textbf{Search Space} \\
       \midrule 
       \# Layers of Aggregation in QATGE & \{1, \textbf{2}\} \\
       Embedding Size & \{100, 200, \textbf{300}\}     \\
       $\gamma$ Initialization & \{0.1, \textbf{0.2}, 0.3\} \\
       $\beta$ Initialization & \{\textbf{0.1}, 0.2, 0.3\} \\
      \bottomrule 
    \end{tabular}
    }
    \caption{Hyperparameter searching strategy. Optimal hyperparameters are marked in bold. The best hyperparameter settings of both datasets are the same.}\label{tab: hyperparameter search}
\end{table}
\begin{table}[t] 
    \begin{center}
    \resizebox{0.75\columnwidth}{!}{
    \begin{tabular}{lcc} 
      \toprule 
     \multicolumn{1}{l}{Datasets} & \multicolumn{1}{c}{\textbf{YAGO-hy}} & \multicolumn{1}{c}{\textbf{Wiki-hy}} \\
      \midrule 
      Model & Training Time  & Training Time \\
       \midrule 
       HypeTKG & 37.53h  & 48.32h \\
       \midrule 
       HypeTKG$^\psi$ & 40.06h  & 51.72h \\ 
      \bottomrule 
    \end{tabular} 
    }
    \end{center}
    \caption{Training time.}
    \label{tab: train time}
\end{table}
\begin{table}[h] 
    \begin{center}
    \resizebox{0.75\columnwidth}{!}{
    \begin{tabular}{lcc} 
      \toprule 
     \multicolumn{1}{l}{Datasets} & \multicolumn{1}{c}{\textbf{YAGO-hy}} & \multicolumn{1}{c}{\textbf{Wiki-hy}} \\
      \midrule 
      Model & GPU Memory  & GPU Memory \\
       \midrule 
       HypeTKG & 9,514MB  & 30,858MB \\
       \midrule 
       HypeTKG$^\psi$ & 15,422MB  & 43,976MB \\ 
      \bottomrule 
    \end{tabular} 
    }
    \end{center}
    \caption{GPU memory usage.}
    \label{tab: memory}
\end{table}
\begin{table}[h] 
    \begin{center}
    \resizebox{0.75\columnwidth}{!}{
    \begin{tabular}{lcc} 
      \toprule 
     \multicolumn{1}{l}{Datasets} & \multicolumn{1}{c}{\textbf{YAGO-hy}} & \multicolumn{1}{c}{\textbf{Wiki-hy}} \\
      \midrule 
      Model & $\#$ Param & $\#$ Param\\
        \midrule 
       HypeTKG & 10,830,222  & 11,028,690 \\ 
       \midrule 
       HypeTKG$^\psi$ & 13,075,246  & 13,274,314 \\ 
      \bottomrule 
    \end{tabular} 
    }
    \end{center}
    \caption{Number of parameters.}
    \label{tab: num_param}
\end{table}

For baselines, we use the official open-sourced implementations of the following baseline methods, i.e., DE-SimplE\footnote{https://github.com/BorealisAI/de-simple}, TeRo\footnote{https://github.com/soledad921/ATISE}, T-GAP\footnote{https://github.com/jaehunjung1/T-GAP}, BoxTE\footnote{https://github.com/JohannesMessner/BoxTE}, TARGCN\footnote{https://github.com/ZifengDing/TARGCN}, TeAST\footnote{https://github.com/dellixx/TeAST}, HGE\footnote{https://github.com/NacyNiko/HGE}, HINGE\footnote{https://github.com/eXascaleInfolab/HINGE\_code}, HypE\footnote{https://github.com/ServiceNow/HypE}, StarE\footnote{https://github.com/migalkin/StarE}, GRAN\footnote{https://github.com/lrjconan/GRAN}, HyConvE\footnote{https://github.com/CarllllWang/HyConvE/tree/master}, ShrinkE\footnote{https://github.com/xiongbo010/ShrinkE} and HyNT\footnote{https://github.com/bdi-lab/HyNT}. For NaLP-Fix, we use its faster implementation in the repository of HINGE. We use the default hyperparameters of all baselines for HTKG LP.

\section{Can TI Knowledge Improve Reasoning over Traditional TKGs?}
\label{app: ti for tkg}
To answer this question, we also enable Variant A (introduced in Sec. \ref{sec: ablation} Ablation Study) to use TI facts and develop Variant A$^\psi$. Since Variant A and A$^\psi$ do not model qualifiers, letting them perform HTKG LP equals doing LP over quadruple-based traditional TKGs. We report Variant A$^\psi$'s LP results in Table \ref{tab: static on TKGs}. By comparing them with Table \ref{tab: proportion}, we find that our TI knowledge modeling components can also effectively enhance reasoning over traditional TKGs.
\begin{table}[t]
    \centering
    \resizebox{\columnwidth}{!}{
    \large\begin{tabular}{@{}lcccccccc@{}}
\toprule
        \textbf{Datasets} & \multicolumn{3}{c}{\textbf{WiKi-hy}} &  \multicolumn{3}{c}{\textbf{YAGO-hy}} \\
\cmidrule(lr){2-4} \cmidrule(lr){5-7}
        \textbf{Model} & MRR & Hits@1 & Hits@10 & MRR & Hits@1  & Hits@10 \\
\midrule 
        Variant A$^\psi$
        & 0.660	& 0.587	& 0.791
        & 0.818	& 0.797	& 0.855
        \\
\bottomrule
    \end{tabular}
    }
    \caption{TKG LP results with time-invariant knowledge.}
\label{tab: static on TKGs}

\end{table}

\section{Impact of Qualifier-Augmented Fact Proportion.}
\label{app: proportion}
To better quantify HypeTKG's power in learning qualifiers, we sample several datasets from Wiki-hy and YAGO-hy with different proportions of facts equipped with qualifiers. We experiment HypeTKG and its variants on these new datasets. 
\paragraph{(100)/(66/(33) Dataset Construction}
We take Wiki-hy as example. We first pick out all the facts, where each of them has at least one qualifier, from Wiki-hy and construct Wiki-hy (100). We call it Wiki-hy (100) because 100\% of its facts are equipped with qualifiers. Next, we keep Wiki-hy (100) and randomly sample an extra number of facts without any qualifier from the original Wiki-hy. We add these facts into Wiki-hy (100) until the proportion of the facts equipped with qualifiers reaches 66\%. We call this new dataset Wiki-hy (66). Similarly, we further expand Wiki-hy (66) to Wiki-hy (33). YAGO-hy (100)/(66)/(33) follows the same policy.
During the process of sampling extra quadruple-based facts, we put each sampled fact to the same set where it comes from. For example, when we construct Wiki-hy (66), we keep Wiki-hy (100) unchanged and further sample quadruple-based facts from Wiki-hy. If a fact is sampled from the training set of Wiki-hy, then it will be put into the training set of Wiki-hy (66). For YAGO-hy, we construct YAGO-hy (100)/(66)/(33) in the same way. We keep the data example proportions of train/valid/test sets in Wiki-hy (100)/(66)/(33) same as the ones in Wiki-hy. YAGO-hy (100)/(66)/(33) follows the same policy. 
Table \ref{tab: proportion stat} shows the dataset statistics of (100)/(66)/(33) datasets used to study the impact of qualifier-augmented fact proportion. As more quadruple-based facts are added, e.g. from (100) to (66), $|\mathcal{E}_\text{pri}|$/$|\mathcal{R}_\text{pri}|$ grows and some entities/relations only existed in qualifiers will appear in primary quadruples, leading to smaller $|\mathcal{E}_{\text{Qual}}|$/$|\mathcal{R}_{\text{Qual}}|$. This does not mean that (100)/(66)/(33) datasets share different pools of qualifier-augmented facts.
Note that the proportions of facts with at least one qualifier in the original Wiki-hy and YAGO-hy are 9.59\% and 6.98\% (Table \ref{tab: data}), respectively, which are much smaller than 33\%. 
\begin{table}[htbp]
    \centering
    \resizebox{\columnwidth}{!}{
\begin{tabular}{c c c c c c c c c}
\toprule
Dataset&$N_\textrm{train}$&$N_\textrm{valid}$&$N_\textrm{test}$&$|\mathcal{E}_\text{pri}|$&$|\mathcal{E}_{\text{Qual}}|$&$|\mathcal{R}_{\text{pri}}|$&$|\mathcal{R}_{\text{Qual}}|$&$|\mathcal{T}|$\\ 
\midrule
Wiki-hy(100)  & $21,210$ & $2,764$ & $2,696$ & $3,392$ & $1,648$ & $25$ & $49$ & $507$\\ 
Wiki-hy(66)  & $31,815$ & $4,146$ & $4,044$ & $8,786$ & $1,643$ & $58$ & $47$ & $507$\\ 
Wiki-hy(33)  & $63,630$ & $8,292$ & $8,088$ & $10,656$ & $1,642$ & $72$ & $46$ & $507$\\ 
YAGO-hy(100)  & $7,232$ & $1,530$ & $1,452$ & $1,739$ & $414$ & $9$ & $33$ & $187$\\ 
YAGO-hy(66)  & $10,848$ & $2,295$ & $2,178$ & $4,844$ & $392$ & $10$ & $33$ & $188$\\ 
YAGO-hy(33)  & $21,696$ & $4,590$ & $4,356$ & $7,339$ & $378$ & $10$ & $33$ & $188$\\ 
\bottomrule
\end{tabular}
}
\caption{(100)/(66)/(33) dataset statistics.}
\label{tab: proportion stat}
\end{table}
\paragraph{Experiments}
We report the performance of HypeTKG and its first three variants on all created datasets in Table \ref{tab: proportion wiki app} and \ref{tab: proportion yago app}. Regardless of the proportion of qualifier-augmented facts, we have two findings: (1) HypeTKG and Variant B \& C benefit from qualifiers on all datasets, confirming the importance of learning qualifiers for reasoning hyper-relational temporal facts.
(2) Variant B \& C constantly underperform HypeTKG on all datasets, proving the effectiveness of both qualifier modeling components.
Note that (100)/(66)/(33) datasets have different data distributions as the original datasets.  
Therefore, it is not meaningful to directly compare each model variant's performance among them (e.g., compare Variant A across Wiki-hy (100)/(66)/(33)). Our findings are based on different variants' performance on the same dataset (e.g., compare Variant A, B, C and HypeTKG on Wiki-hy(100)).
\begin{table*}[t]
    \centering
    \resizebox{0.95\textwidth}{!}{
    \large\begin{tabular}{@{}l|ccc|ccccccccc@{}}
\toprule
        & \multicolumn{3}{c|}{\textbf{Setting}}  &  \multicolumn{3}{c}{\textbf{Wiki-hy (33)}} &  \multicolumn{3}{c}{\textbf{Wiki-hy (66)}} &  \multicolumn{3}{c}{\textbf{Wiki-hy (100)}}\\
\cmidrule(lr){2-4} \cmidrule(lr){5-7} \cmidrule(lr){8-10} \cmidrule(lr){11-13} 
        \textbf{Model} & Time & Q Att &  Match & MRR & H@1 & H@10 & MRR & H@1 & H@10 & MRR & H@1 & H@10 \\
\midrule 
        Variant A  & \ding{51} & \ding{55}& \ding{55}
        & 0.499 & 0.420 & 0.624
        & 0.522 & 0.457 & 0.622
        & 0.629 & 0.562 & 0.739
         \\
        Variant B  & \ding{51} & \ding{55} & \ding{51}
        & 0.520 & 0.462 & 0.626 
        & 0.570 & 0.528 & 0.638
        & 0.669 & 0.622 & 0.749
         \\
        Variant C & \ding{51} & \ding{51} & \ding{55}
        & 0.519 & 0.461 & 0.622 
        & 0.567 & 0.524 & 0.639
        & 0.662 & 0.607 & 0.749
         \\
        HypeTKG & \ding{51} & \ding{51}& \ding{51}
        & \textbf{0.546} & \textbf{0.492} & \textbf{0.642} 
        & \textbf{0.573} & \textbf{0.531} & \textbf{0.642}
        & \textbf{0.682} & \textbf{0.640} & \textbf{0.750}
        \\
\bottomrule
    \end{tabular}
    }
    \caption{Study of qualifier-augmented fact proportion on Wiki-hy.
    }
    \label{tab: proportion wiki app}
\end{table*}
\begin{table*}[t]
    \centering
    \resizebox{0.95\textwidth}{!}{
    \large\begin{tabular}{@{}l|ccc|ccccccccc@{}}
\toprule
        & \multicolumn{3}{c|}{\textbf{Setting}} &  \multicolumn{3}{c}{\textbf{YAGO-hy (33)}} &  \multicolumn{3}{c}{\textbf{YAGO-hy (66)}} &  \multicolumn{3}{c}{\textbf{YAGO-hy (100)}}\\
\cmidrule(lr){2-4} \cmidrule(lr){5-7} \cmidrule(lr){8-10} \cmidrule(lr){11-13} 
        \textbf{Model} & Time & Q Att & Q Match & MRR & H@1 & H@10 & MRR & H@1 & H@10 & MRR & H@1 & H@10 \\
\midrule 
        Variant A  & \ding{51} & \ding{55}& \ding{55}
        & 0.650 & 0.624 & 0.694 
        & 0.574 & 0.531 & 0.644
        &0.593 & 0.576 & 0.622
         \\
        Variant B  & \ding{51} & \ding{55} & \ding{51} 
        & 0.692 & 0.673 & 0.701 
        & 0.640 & 0.629 & 0.668
        &0.685 & 0.672 & 0.706
         \\
        Variant C & \ding{51} & \ding{51} & \ding{55}
        & 0.687 & 0.669 & 0.700 
        & 0.638 & 0.625 & 0.667
        &0.683 & 0.670 & 0.705
         \\
        HypeTKG & \ding{51} & \ding{51}& \ding{51}
        & \textbf{0.696} & \textbf{0.678} & \textbf{0.703} 
        & \textbf{0.645} & \textbf{0.632} & \textbf{0.669} 
        & \textbf{0.688} & \textbf{0.676} & \textbf{0.712}
        \\
\bottomrule
    \end{tabular}
    }
\caption{Study of qualifier-augmented fact proportion on YAGO-hy.
    }
    \label{tab: proportion yago app}
\end{table*}

\begin{table*}[htbp]
\small
    \centering
    \resizebox{\textwidth}{!}{
\begin{tabular}{c| c| c| c| c} 
\toprule
\textbf{Case}&\textbf{Query}&\textbf{Answer}&\textbf{Subject-Related Qualifiers}&\textbf{Attention Score}
\\ 
\midrule
\multirow{3}{*}{A1}&\multirow{3}{*}{$((\textit{Andrey Kolmogorov}, \textit{award received}, ?, 1941), \emptyset)$}  &	\multirow{3}{*}{$\textit{USSR State Prize}$} & $(\textit{country of citizenship}, \textit{Soviet Union})$ & $9.39e^{-1}$  \\

&&& $(\textit{field of work}, \textit{mathematics})$ & $6.09e^{-2}$\\
&&& $(\textit{country}, \textit{Soviet Union})$ & $2.61e^{-10}$\\
\midrule
\multirow{3}{*}{A2}&\multirow{2}{*}{$((\textit{Andrey Kolmogorov}, \textit{place of death}, ?, 1987),$}  &	\multirow{3}{*}{$\textit{Moscow}$} & $(\textit{country of citizenship}, \textit{Soviet Union})$ & $0.99$  \\

&&& $(\textit{field of work}, \textit{mathematics})$ & $1.64e^{-21}$\\
&$\{(\textit{country}, \textit{Soviet Union})\})$&& $(\textit{country}, \textit{Soviet Union})$ & $5.00e^{-22}$\\
\midrule
\multirow{2}{*}{A3}&\multirow{2}{*}{$((\textit{Wernher von Braun}, \textit{academic degree}, ?, 1934), \emptyset)$}  &	\multirow{2}{*}{$\textit{Doctor of Philosophy}$} & $(\textit{academic degree}, \textit{Doctor of Philosophy})$ & $0.99$  \\

&&& $(\textit{academic major}, \textit{physics
})$ & $6.00e^{-10}$\\
\bottomrule
\end{tabular}
}
\caption{Case study A: cases for studying qualifier matcher.}
\label{tab: case app}
\end{table*}
\begin{table*}[htbp]
\small
    \centering
    \resizebox{\textwidth}{!}{
\begin{tabular}{c| c| c| c| c} 
\toprule
\textbf{Case}&\textbf{Query}&\textbf{Prediction w. TI}&\textbf{Prediction w.o. TI} & \textbf{Related TI Facts}
\\ 
\midrule
\multirow{3}{*}{B1}&\multirow{3}{*}{$((\textit{Pisa}, \textit{country}, ?, 1860), \emptyset)$}  &	\multirow{3}{*}{$\textit{Kingdom of Sardinia}$} & \multirow{3}{*}{$\textit{Kingdom of Prussia}$}  & $(\textit{Pisa}, \textit{official language}, \textit{Italian})$\\
&&&& $(\textit{Kingdom of Sardinia}, \textit{official language}, \textit{Italian})$\\
&&&& $(\textit{Kingdom of Prussia}, \textit{official language}, \textit{German})$\\
\midrule
\multirow{2}{*}{B2}&\multirow{2}{*}{$((\textit{AK}, \textit{place of birth}, ?, 1903), \{(\textit{country}, \textit{Russian Empire})\})$}  &	\multirow{2}{*}{$\textit{Tbilisi}$} & \multirow{2}{*}{$\textit{Moscow}$} & $(\textit{AK}, \textit{native language}, \textit{Georgian})$\\
&&&& $(\textit{Tbilisi}, \textit{official language}, \textit{Georgian})$\\
\bottomrule
\end{tabular}
}
\caption{Case study B: cases for studying the effectiveness of TI relational knowledge. Prediction w./w.o. TI means the prediction result with/without using time-invariant facts. \textit{AK} is the abbreviation of the entity \textit{Aram Khachaturian}.}
\label{tab: case2 app}
\end{table*}
\section{Case Study Details}
\label{app: case}
\paragraph{A: Effectiveness of Qualifier Matcher}
\label{app: case}
We give an insight of how our qualifier matcher improves HTKG reasoning with three cases (Table \ref{tab: case}). HypeTKG ranks the ground truth missing entities in these cases as top 1 and achieves optimal prediction. As discussed in Sec. \ref{sec: decoder}, we learn a global qualifier feature in the qualifier matcher by considering the contribution of all the existing qualifiers related to the subject entity of the LP query. Each qualifier is assigned an attention score $\eta_l$ indicating its contribution. Note that numerous queries are derived from the facts that are without any qualifier. For example, in Case A1, no qualifier is provided in predicting which reward did \textit{Andrey Kolmogorov} receive in 1941 (Case A1 and A2 are taken from YAGO-hy). HypeTKG extracts all the qualifiers related to \textit{Andrey Kolmogorov} from other facts in YAGO-hy and computes the global qualifier feature based on them. We find that it assigns a great attention score to the qualifier $(\textit{country of citizenship}, \textit{Soviet Union})$ and this qualifier can directly be taken as a hint to predict the ground truth missing entity \textit{USSR State Prize} since USSR is also interpreted as Soviet Union. We also find that $(\textit{field of work}, \textit{mathematics})$ is also dominant in the global qualifier feature. This is also reasonable because Andrey Kolmogorov was a mathematician and he was awarded USSR State Prize of mathematics in 1941. Compared with these two qualifiers, the last qualifier, i.e., $\{(\textit{country}, \textit{Soviet Union})\})$, is not so important in prediction, and thus is assigned a low attention score by HypeTKG. Case A1 implies that to reason the facts without qualifiers, i.e., quadruple-based facts, our qualifier matcher can find the clues from the subject-related qualifiers existing in other hyper-relational facts and support prediction. In Case A2, we find that the qualifier matcher focuses more on the qualifiers from other facts but not the one from the query. Note that the query qualifiers have been explicitly modeled with a query-specific qualifier feature $\mathbf{h}^\text{que}_\text{Qual}$ before computing the global qualifier feature. This indicates that our qualifier matcher can maximally extract important information from the extra qualifiers rather than only focusing on the query qualifiers, enabling efficient information fusion. 
Case A3 is taken from Wiki-hy. Since qualifier relations and primary relations have intersection, some extra subject-related qualifiers from other HTKG facts can directly indicate the answers to the queries. In Case A3, we observe that HypeTKG manages to recognize such qualifiers to improve prediction. This further proves that our qualifier matcher is able to help capture the correlation between qualifiers and temporal validity. To summarize, our qualifier matcher achieves reasoning enhancement by efficiently utilizing additional information from the extra qualifiers related to the query subject.
\paragraph{B: Effectiveness of TI Knowledge}
\label{app: case2}
We demonstrate how TI relational knowledge enhances HTKG reasoning with two cases (Table \ref{tab: case2}). In both cases, HypeTKG achieves optimal prediction (ranks ground truth answers as top 1) by leveraging TI knowledge, and makes mistakes without considering it. Case B1 is taken from Wiki-hy. In B1, HypeTKG predicts the false answer $\textit{Kingdom of Prussia}$ without the support of TI facts. However, after considering them, HypeTKG manages to make accurate prediction because $\textit{Pisa}$ should share the same official language with the country that contains it. Case B2 is taken from YAGO-hy. In B2, since both \textit{Tbilisi} and \textit{Moscow} belonged to \textit{Russian Empire} in 1903, it is hard for HypeTKG to distinguish them during prediction without any further information. However, by knowing that \textit{Aram Khachaturian}'s native language is same as the official language of \textit{Tbilisi}, i.e., Georgian, HypeTKG can exclude the influence of \textit{Moscow} because people speak Russian there. The presented cases illustrate how our model better reasons HTKGs with TI knowledge.

\section{Related Work Details}
\label{app: related work}
\subsection{Traditional KG \& TKG Reasoning}
Extensive research has been conducted for KG reasoning. A series of works \cite{DBLP:conf/nips/BordesUGWY13,DBLP:conf/icml/TrouillonWRGB16,DBLP:conf/iclr/SunDNT19,DBLP:conf/nips/0007TYL19,DBLP:conf/aaai/CaoX0CH21,DBLP:conf/kdd/XiongZNXP0S22,xiong2022faithful,DBLP:conf/acl/NayyeriXMAA0S23} designs KG score functions that compute plausibility scores of triple-based KG facts, while another line of works \cite{DBLP:conf/esws/SchlichtkrullKB18,DBLP:conf/iclr/VashishthSNT20} incorporates neural-based modules, e.g., graph neural network (GNN) \cite{DBLP:conf/iclr/KipfW17}, into score functions for learning better representations. 
On top of the existing KG score functions, some recent works develop time-aware score functions \cite{DBLP:conf/www/LeblayC18,DBLP:conf/coling/XuNAYL20,DBLP:conf/aaai/GoelKBP20,DBLP:journals/kbs/ShaoZYTCL22,DBLP:conf/aaai/MessnerAC22,DBLP:conf/acl/LiSG23,DBLP:conf/cikm/XiongNDC23,DBLP:conf/aaai/PanNLS24} that further model time information for reasoning over traditional TKGs. Another group of TKG reasoning methods employ neural structures. Some of them \cite{DBLP:conf/emnlp/JinQJR20,DBLP:conf/emnlp/WuCCH20,DBLP:conf/emnlp/HanDMGT21,DBLP:conf/aaai/ZhuCFCZ21,DBLP:conf/sigir/LiJLGGSWC21,DBLP:conf/ijcai/LiS022,DBLP:conf/icde/Liu0X0023,DBLP:conf/naacl/DingCWMLXT24} achieve temporal reasoning by first learning the entity and relation representations of each timestamp with GNNs and then using recurrent neural structures, e.g., LSTM \cite{article}, to compute time-aware representations. Other methods \cite{DBLP:conf/kdd/JungJK21,DBLP:conf/iclr/HanCMT21,ding2022a} develop time-aware relational graph encoders that directly perform graph aggregation based on the temporal facts sampled from different time. There are two settings in TKG LP, i.e., interpolation and extrapolation. In extrapolation, to predict a fact happening at time $t$, models can only observe previous TKG facts before $t$, while such restriction is not imposed in interpolation. Among the above mentioned works, \cite{DBLP:conf/www/LeblayC18,DBLP:conf/coling/XuNAYL20,DBLP:conf/aaai/GoelKBP20,DBLP:journals/kbs/ShaoZYTCL22,DBLP:conf/aaai/MessnerAC22,DBLP:conf/emnlp/WuCCH20,DBLP:conf/kdd/JungJK21,ding2022a,DBLP:conf/acl/LiSG23,DBLP:conf/cikm/XiongNDC23,DBLP:conf/aaai/PanNLS24} are for interpolation and \cite{DBLP:conf/emnlp/JinQJR20,DBLP:conf/emnlp/HanDMGT21,DBLP:conf/aaai/ZhuCFCZ21,DBLP:conf/sigir/LiJLGGSWC21,DBLP:conf/iclr/HanCMT21,DBLP:conf/ijcai/LiS022,DBLP:conf/icde/Liu0X0023,DBLP:conf/naacl/DingCWMLXT24} are for extrapolation. Traditional TKG reasoning methods cannot optimally reason over HTKG facts because they are unable to model qualifiers. In our work, we only focus on the interpolated LP on HTKGs and leave extrapolation for future work.

\subsection{Hyper-Relational KG Reasoning}
Mainstream HKG reasoning methods can be categorized into three types.
The first type of works \cite{DBLP:conf/www/ZhangLMM18,DBLP:conf/www/0016Y020,DBLP:conf/ijcai/FatemiTV020,DBLP:conf/www/DiYC21,DBLP:conf/www/WangWLC023} treats each hyper-relational fact as an $n$-ary fact represented with an $n$-tuple: $r_\text{abs}(e_1,e_2, ..., e_n)$, where $n$ is the non-negative arity of an abstract relation $r_\text{abs}$ representing the number of entities involved within $r_\text{abs}$ and $e_1, ..., e_n$ are the entities appearing in this $n$-ary fact. RAE \cite{DBLP:conf/www/ZhangLMM18} generalizes traditional KG reasoning method TransH \cite{DBLP:conf/aaai/WangZFC14} to reasoning $n$-ary facts and improves performance by considering the relatedness among entities. Similarly, HypE \cite{DBLP:conf/ijcai/FatemiTV020} and GETD \cite{DBLP:conf/www/0016Y020} derive the $n$-ary fact reasoning models by modifying traditional KG score functions SimplE \cite{DBLP:conf/nips/Kazemi018} and TuckER \cite{DBLP:conf/emnlp/BalazevicAH19}, respectively. S2S \cite{DBLP:conf/www/DiYC21} improves GETD by enabling reasoning over mixed-arity facts. HyConvE \cite{DBLP:conf/www/WangWLC023} employs convolutional neural networks to perform 3D convolution capturing the deep interactions of entities and relations. Although these methods show strong effectiveness, the way of treating HKG facts as $n$-ary facts naturally loses the semantics of the original KG relations and would lead to a combinatorial explosion of relation types \cite{DBLP:conf/emnlp/GalkinTMUL20}.
The second type of works \cite{DBLP:conf/www/LiuYL21,DBLP:journals/tkde/GuanJGWC23} transforms each hyper-relational fact into a set of key-value pairs: $\{(r_i:e_i)\}_{i=1}^n$. RAM \cite{DBLP:conf/www/LiuYL21} introduces a role learning paradigm that models both the relatedness among different entity roles as well as the role-entity compatibility. NaLP \cite{DBLP:journals/tkde/GuanJGWC23} captures the relatedness among all
the $r_i:e_i$ pairs by using neural networks. Formulating hyper-relational facts into solely key-value pairs would also cause a problem. The relations from the primary fact triples and qualifiers cannot be fully distinguished, and the semantic difference among them is ignored \cite{DBLP:conf/emnlp/GalkinTMUL20}.
To overcome the problems incurred in first two types of methods, recently, some works \cite{DBLP:conf/acl/GuanJGWC20,DBLP:conf/www/RossoYC20,DBLP:conf/emnlp/GalkinTMUL20,DBLP:conf/acl/WangWLZ21,DBLP:conf/acl/XiongNPS23} formulate each hyper-relational fact into a primary triple with a set of key-value qualifier pairs: $\{((s,r,o),\{(r_{q_i}, e_{q_i})\}_{i=1}^n)\}$. NeuInfer \cite{DBLP:conf/acl/GuanJGWC20} uses fully-connected neural networks to separately model each primary triple and its qualifiers. HINGE \cite{DBLP:conf/www/RossoYC20} adopts a convolutional framework that is iteratively applied on the qualifiers for information fusion. StarE \cite{DBLP:conf/emnlp/GalkinTMUL20} develops a qualifier-aware GNN which allows jointly modeling an arbitrary number of qualifiers with the primary triple relation. GRAN \cite{DBLP:conf/acl/WangWLZ21} models HKGs with edge-biased fully-connected attention. It uses separate edge biases for the relations in the primary triples and qualifiers to distinguish their semantic difference. ShrinkE \cite{DBLP:conf/acl/XiongNPS23} models each primary triple as a spatial-functional transformation from the primary subject to a relation-specific box and lets qualifiers shrink the box to narrow down the possible answer set. Based on them, NestE \cite{DBLP:conf/aaai/XiongNLWPS24} considers nested relational structure where a fact is composed of other facts. It focuses on nested facts that can be viewed as a generalization of hyper-relational facts.

A recent work \cite{DBLP:conf/emnlp/HouJL0GZGC23} proposes a new type of TKG, i.e., n-tuple TKG (N-TKG), where each hyper-relational fact is represented with an n-tuple: 
$(r, \{\rho_i: e_i\}_{i=1}^n, t)$. $n$ and $t$ are the arity and the timestamp of the fact, respectively. $\rho_i$ is the labeled role of the entity $e_i$. $r$ denotes fact type. Compared with HTKG, N-TKG has limitation: HTKGs explicitly separate primary facts with additional qualifiers, while N-TKGs mix all the entities from the primary facts and qualifiers and are unable to fully emphasize the importance of primary facts.
Besides, N-TKGs pair each entity with a labeled role. A large proportion of roles are not directly extracted from the associated KBs and are manually created depending on the fact type (e.g., the proposed NICE dataset in \cite{DBLP:conf/emnlp/HouJL0GZGC23}). In our work, qualifiers are directly taken from the Wikidata KB, which guarantees that all the additional information conforms to the original KB and requires no further effort of manual labeling. Another drawback of \cite{DBLP:conf/emnlp/HouJL0GZGC23} is that the proposed NICE N-TKG dataset  in this work is based on ICEWS KB. As discussed in App. \ref{app: why not icews}, using ICEWS for constructing hyper-relational KGs does not fully align to the motivation of introducing qualifiers into traditional TKGs. Our proposed HTKGs are both based on Wikidata KB, which is much more meaningful.
To achieve extrapolated LP over N-TKGs, \cite{DBLP:conf/emnlp/HouJL0GZGC23} develops a model called NE-Net that jointly learns from historical temporal information and entity roles. NE-Net performs well on N-TKG extrapolation, but it is not optimal for interpolation over hyper-relational facts because it is unable to encode the graph information after the timestamp of each LP query. Our proposed HTKG reasoning model HypeTKG is able to capture the temporal factual information along the whole timeline of HTKGs, serving as a more reasonable method for interpolated LP.

\end{document}